\documentclass[journal,10pt]{IEEEtran} % refer to the author guide from IEEE

\ifCLASSOPTIONcompsoc
   \usepackage[nocompress]{cite}
\else
   \usepackage{cite}
\fi

\usepackage{times,amsmath,amssymb,amsfonts,algorithm,epsfig,subfigure,cite}
\usepackage{color} % for revision only
\usepackage[algo2e]{algorithm2e}
\usepackage{multirow}
\usepackage{makecell}
\usepackage{color}
\usepackage{bm}

\usepackage[colorlinks,
            linkcolor=blue,
            anchorcolor=blue,
            citecolor=blue,
            ]{hyperref}

\newcommand{\PreserveBackslash}[1]{\let\temp=\\#1\let\\=\temp}
\newcolumntype{C}[1]{>{\PreserveBackslash\centering}p{#1}}
\newcolumntype{R}[1]{>{\PreserveBackslash\raggedleft}p{#1}}
\newcolumntype{L}[1]{>{\PreserveBackslash\raggedright}p{#1}}

\usepackage{booktabs}
\def\X{{\mathbf{X}}}
\def\U{{\mathbf{U}}}
\def\V{{\mathbf{V}}}

\def\A{{\mathbf{A}}}

\def\G{{\mathbf{G}}}

\def\E{{\mathbf{E}}}

\def\S{{\mathbf{S}}}

\def\Y{{\mathbf{Y}}}
\usepackage{algorithm}
\usepackage{algorithmic}
\renewcommand{\arraystretch}{1.12}
\usepackage{array}
\newcolumntype{C}[1]{>{\PreserveBackslash\centering}p{#1}}
\newtheorem{theorem}{Theorem}

\newtheorem{definition}{Definition}

\newtheorem{proof}{Proof}

\begin{document}
	%****************************************************************************
	% Title
%	\title{	Unknown Noise Removal for Hyperspectral Image via Total Variation on Representative Coefficient}
	
%	\title{Hyperspectral Image Denoising via Total Variation Regularization on Representative Coefficient}
	
%	\title{Representative- Coefficient-total-variation-Regularized for Mixed Noise Removal in Hyperspectral Image}
	
	\title{Fast Noise Removal in Hyperspectral Images via Representative Coefficient Total Variation}
%	\title{Hyperspectral Mixed Noise Removal via Weighted Representative Coefficient Total Variation Regularization}

	%****************************************************************************
	\author{Jiangjun~Peng,
	 Hailin~Wang,
	 Xiangyong~Cao,~\IEEEmembership{Member,~IEEE},
	 Xinlin~Liu,
	 Xiangyu~Rui,
	 and
	 Deyu~Meng,~\IEEEmembership{Member,~IEEE}
 \thanks{Jiangjun~Peng, Hailin~Wang (\emph{co-first author}), Xiangyu~Rui, and Deyu~ Meng are with School of Mathematics and Statistics and Ministry of Education Key Lab of Intelligent Networks and Network Security, Xi'an Jiaotong University, Xi'an 710049, Shaan'xi, China. Email: andrew.pengjj@gmail.com, wanghailin97@163.com, xyrui.aca@gmail.com, dymeng@mail.xjtu.edu.cn.}
\thanks{Xiangyong Cao is with School of Computer Science and Technology and Ministry of Education Key Lab For Intelligent Networks and Network
Security, Xi’an Jiaotong University, Xi’an 710049, China.
E-mail:caoxiangyong@xjtu.edu.cn.}
\thanks{Xinling Liu is with the School of Mathematics and Statistics, Southwest University, Chongqing 400715, China, and also with School of Mathematics and Information, China West Normal University, Nanchong 637002, China. E-mail: fsliuxl@163.com.}
}
	%\markboth{submitted to IEEE Transactions on xxx,~Vol.~XX, No.~XX, XXX~2020}{}
	%****************************************************************************
	
	\maketitle
	%****************************************************************************
	%
	\begin{abstract}
Mining structural priors in data is a widely recognized technique for hyperspectral image (HSI) denoising tasks, whose typical ways include model-based methods and data-based methods. The model-based methods have good generalization ability, while the runtime cannot meet the fast processing requirements of the practical situations due to the large size of an HSI data $ \X \in \mathbb{R}^{MN\times B}$. For the data-based methods, they perform very fast on new test data once they have been trained. However, their generalization ability is always insufficient. In this paper, we propose a fast model-based HSI denoising approach. Specifically, we propose a novel regularizer named Representative Coefficient Total Variation (RCTV) to simultaneously characterize the low rank and local smooth properties. The RCTV regularizer is proposed based on the observation that the representative coefficient matrix $\U\in\mathbb{R}^{MN\times R} (R\ll B)$ obtained by orthogonally transforming the original HSI $\X$ can inherit the strong local-smooth prior of $\X$. Since $R/B$ is very small, the HSI denoising model based on the RCTV regularizer has lower time complexity. Additionally, we find that the representative coefficient matrix $\U$ is robust to noise, and thus the RCTV regularizer can somewhat promote the robustness of the HSI denoising model. Extensive experiments on mixed noise removal demonstrate the superiority of the proposed method both in denoising performance and denoising speed compared with other state-of-the-art methods. Remarkably, the denoising speed of our proposed method outperforms all the model-based techniques and is comparable with the deep learning-based approaches.
	\end{abstract}
	
	
	\IEEEpeerreviewmaketitle
	
\section{Introduction}
Hyperspectral image (HSI) data are acquired by high spectral resolution sensors, and consist of hundreds of contiguous narrow spectral band images from ultraviolet to infrared wavelengths for the same object \cite{green1998imaging}. Since HSI contains abundant spatial and spectral information, it is thus widely used in various applications, such as object detection \cite{liu2016tensor}, super-resolution \cite{dong2016hyperspectral}, unmixing \cite{yao2019nonconvex}, mineral exploration \cite{goetz2009three} and classification \cite{jia20173, cao2020hyperspectral, zhao2020joint}. However, due to the sensor sensitivity, calibration error, physical mechanism, and weather interference, HSI is unavoidably contaminated by various kinds of noise, such as Gaussian noise, stripe noise, deadline noise, impulse noise, and so on \cite{shen2008map, skauli2011sensor, he2015total, wang2017hyperspectral}. These noise severely degrades the quality of the imagery and limits the performance of the subsequent processing. Therefore, HSI denoising is an important pre-processing step to improve image quality for further downstream tasks.

The current HSI denoising methods can be roughly divided into two categories, namely model-based methods and data-based methods.  Data-based methods mainly learn a supervised deep neural network from carefully collected noisy-clean HSI pairs, and can be tested quickly. Although the data-based methods perform fast on the test data, they incline to be with poor generalization capability in complex test scenarios differing from their training cases, mainly attributed to their strong over-fitting ability to training data \cite{yuan2018hyperspectral, chang2018hsi, cao2021deep}. On the contrary, the unsupervised model-based methods are based on statistical prior modeling in HSI, and thus possess good performance in generalization. However, most of the model-based methods need a relatively long time to implement \cite{he2015total, chang2017hyper, chang2018hsi}, making it challenging  to meet the practical demand in HSI denoising. Therefore, it should be a significant issue for model-based HSI denoising research to accelerate their implementation efficiency on test data like deep learning methods, while maintain their general denoising performance simultaneously. This paper focuses on this goal and attempt to design a fast model-based HSI denoising method.

It is known that almost all model-based HSI denoising methods, such as LRTV \cite{he2015total}, LLRT \cite{chang2017hyper}, LRTDTV \cite{wang2017hyperspectral}, LRMR \cite{zhang2013hyperspectral}, and CTV-RPCA \cite{peng2022exact}, are built by mining the prior of the original HSI data $ \mathcal{X} \in \mathcal{R}^{M\times N \times B} $, where $ M, N $ and $ B $ denote the spatial height, the spatial width, and the number of bands, respectively. Among these priors, low-rank (L) prior along the spectral dimension is the most indispensable one, which usually needs singular value decomposition (SVD) to solve related low-rank optimization problems \cite{candes2011robust}. Besides, HSI data also possesses the local smoothness (LS) and non-local similarity (NLS) priors in the spatial dimensions \cite{he2015total, chang2017hyper}. Generally, the LS prior is encoded by the Total Variation (TV) regularization \cite{rudin1992nonlinear}, and the related TV minimization problem is usually solved by the Fast Fourier Transform \cite{beck2009fast, krishnan2009fast} (FFT). The NLS prior requires Similarity Block Search (SBS) step to get many groups, and then conducts an SVD operator on each group. The specific time complexity of the aforementioned L, LS and NLS priors are given in Table \ref{prior_complexity}. Since $ \mathcal{O}(M^2N^2B n^2/r^4)$ is much larger than $\mathcal{O}(MNB\log(MB))$ and $ \mathcal{O}(MNB^2 ) $ is approximately equal to $ \mathcal{O}(Kpn^2B^2) $, the NLS prior has the highest time complexity. Besides, it can be observed that all the prior terms have a positive correlation with the size of HSI, the SVD computing is very time-consuming, and the TV prior with FFT is relatively efficient. For most advanced/classic model-based models, such as LRTV \cite{he2015total}, LRTDTV \cite{wang2017hyperspectral}, and E3DTV \cite{peng2020enhanced}, they often utilize multiple priors on the original HSI data simultaneously to obtain better visual performance, making their execution speeds very slow since the original data size is quite large, especially on the HSIs with a large band number $ B $, which is often encountered in real applications. 

\begin{table}[!t]
\begin{minipage}[!t]{\columnwidth}
  \renewcommand{\arraystretch}{1.2}
  \caption{The time complexity of  above mentioned priors.}
  \label{prior_complexity}
  \centering
    \vspace{-0.2cm}
  \setlength{\tabcolsep}{2.5pt}{
  \begin{tabular}{c|l|l|l}
	\Xhline{1pt}
	Prior Term & Subject & Algorithm  & Time Complexity \\
	\hline
	\hline
	\textbf{L} & $\mathcal{X} \in \mathbb{R}^{M\times N \times B}$ & SVD & $ \mathcal{O}(MNB^2) $ \\
	\Xhline{0.5pt}
	\textbf{LS} & $\mathcal{X} \in \mathbb{R}^{M\times N \times B}$ & FFT & $ \mathcal{O}(MNB\log(MN)) $ \\
	\Xhline{0.5pt}
	\multirow{2}{*}{\makecell[c]{\textbf{NLS}}} &\multirow{2}{*}{\makecell[c]{$\mathcal{X} \in \mathbb{R}^{M\times 		N \times B}$}}  & SBS\footnote{Specifically, we need to divide the HSI data to obtain lots of patches with the size of $ n\times n \times B $ and then search for similar patches to aggregate them into $ K $ groups, each group containing $ p $ patches. Assuming that the stride of the division is $ r $, then $ r\leq n $ is satisfied, we can get $ MN/r^2 $ patches.} & $ \mathcal{O}( M^2N^2Bn^2/r^4) $ \\
	\cline{3-4}
	& & SVD & $\mathcal{O}(KB^2pn^2)$ \\
	\hline
	\hline
	\textbf{LS} & $\U \in \mathbb{R}^{MN \times R}$ & FFT & $ \mathcal{O}(MNR\log(MN)) $ \\
	\Xhline{1pt}
\end{tabular}}
  \end{minipage}
\\[12pt]
\begin{minipage}[!t]{\columnwidth}
  \renewcommand{\arraystretch}{1.25}
      \vspace{-0.3cm}
  \caption{The time complexity of some classical models and their running times on data with a size of $ 512\times 512\times 31 $.}
  \label{model_complexity}
  \centering
  \setlength{\tabcolsep}{2.5pt}{
    \begin{tabular}{l|l|l|l}
	\Xhline{1pt}
	Models &Prior Terms &Time Complexity  & Time(s) \\
	\hline
	\hline
	LLRT& NLS, L of $ \X $&$\mathcal{O}( M^2N^2Bn^2/r^4+Kpn^2B^2)$& 1593\\
	\Xhline{0.5pt}
	NGmeet & NLS, L of $ \U $&$\mathcal{O}( M^2N^2Rn^2/r^4+(Kpn^2+B)R^2)$& 195.6\\
	\Xhline{0.5pt}
	LRTV\footnote{Since LRTV uses an iterative gradient descent algorithm to solve the TV problem, the time complexity of TV subproblem is related to the number of iteration steps $ T $.} & LS, L of $ \X $&$\mathcal{O}( MNB^2 +TMNB)$& 270.6\\
	\Xhline{0.5pt}
	CTV & LS, L of $ \X $&$\mathcal{O}(MNB\log(MN)+MNB^2)$&67.6\\
	\Xhline{0.5pt}
	RCTV & LS, L of  $ \U $&$\mathcal{O}(MNR\log(MN)+R^2B)$&11.2\\
	\Xhline{1pt}
\end{tabular}}
  \end{minipage}
  \vspace{-0.1cm}
\end{table}

	\begin{figure}[!t]
		\centering
		\includegraphics[scale=0.3]{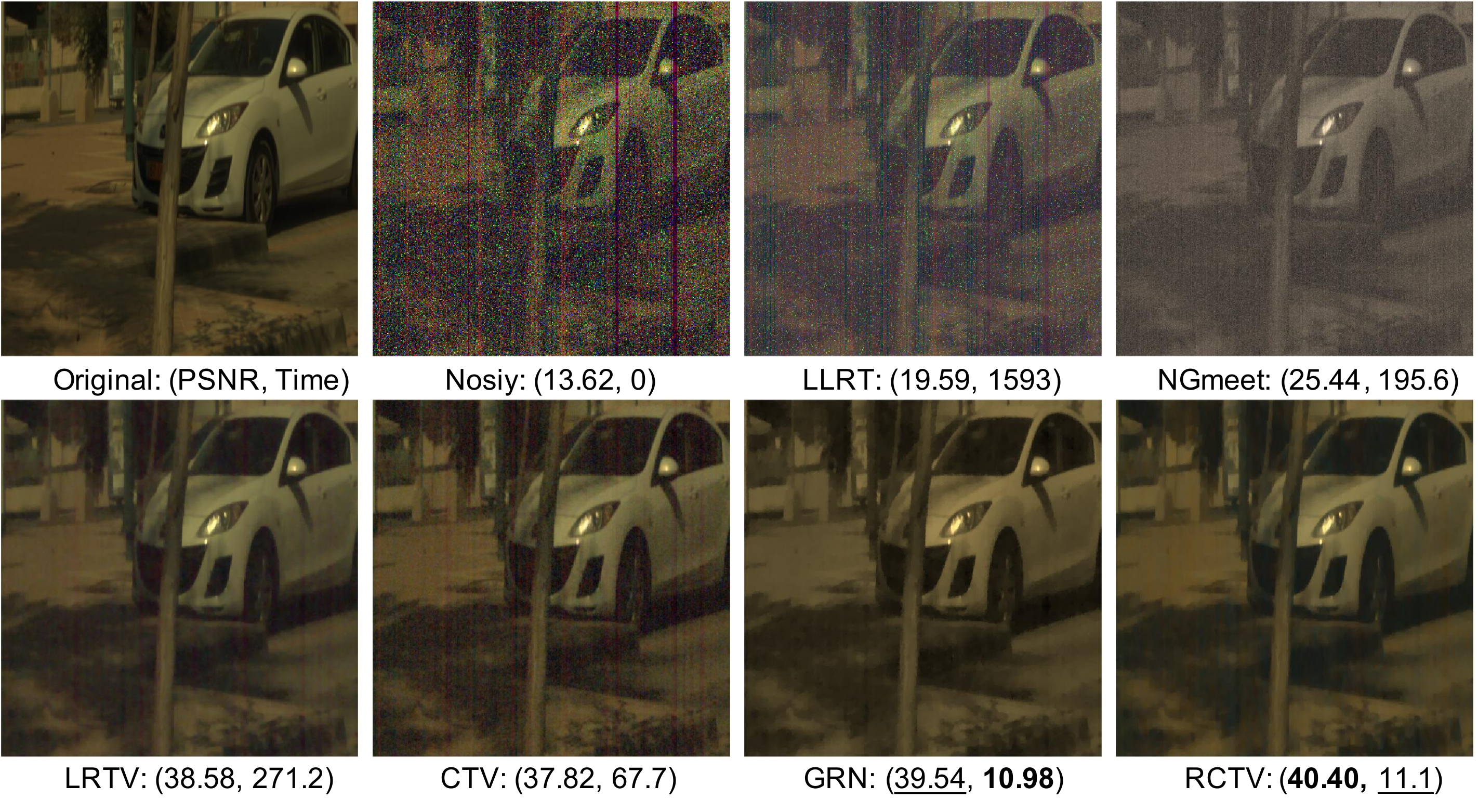}
		\vspace{-4mm}
		\caption{The restoration result of models in Table \ref{model_complexity} on one data in the ICVL dataset.}
		\label{show_introduction}
	\end{figure}

Is it possible that the original data has a factor with a smaller size that inherits the prior information of $ \X $? If this is true,  we can significantly reduce denoising time while guaranteeing denoising performance. Inspired by this motivation, we turn attention to low-rank matrix factorization (LRMF). For an HSI data $ \mathcal{X} \in \mathbb{R}^{M\times N\times B} $ with unfolded matrix $ \X \in \mathbb{R}^{M N\times B} $ along the spectral dimension, suppose the rank of $ \X $ is $R$ (it generally holds that $R\ll B $), and then we have $ X=\U\V^{\textrm{T}} $, where $ \U \in \mathbf{R}^{MN\times R} $, and $ \V \in \mathbf{R}^{B\times R} $ are called the abundance map and the end-member matrix\cite{yao2019nonconvex}, respectively. Mathematically, an abundance map $ \U $ describes the representative coefficients of a row vector in $ \X $ for a given basis matrix $  \V$. Therefore, $ \U $ can also be called the representative coefficient matrix. Such factorization naturally reflects the spectral low-rankness of HSI data by controlling the rank number $ R$, which also has the benefit that the optimization of the LRMF-based model does not need large-scale SVD computation. Further, we first prove in Theorem \ref{theorem_e} that $ \U $ can inherit the spatial structure of $ \X $. Thus, it is naturally hopeful to build a model with low time complexity and high restoration performance based on priors of $ \U $. In addition, we find interestingly that the images corresponding to the intrinsic representative coefficient matrix $ \U $ can better maintain the structure of the original image than the noisy image, which means that the representative coefficient matrix has certain robustness to noise and thus can hopefully obtain better denoising performance. Since LS prior has lower time complexity than NLS prior, we propose RCTV regularization based on LS prior of $ \U $, and its time complexity is also listed in Table \ref{prior_complexity}. As can be seen from Table \ref{prior_complexity}, compared to using the LS prior of the original data $ \X $, the time complexity of using the LS prior of $ \U $ can be greatly reduced.

Low-complexity regularizers tend to lead to low-complexity models. To better illustrate this, we list the time complexity of some classical models in Table \ref{model_complexity} \footnote{Since each model contains some matrix multiplication operations with time complexity $\mathcal{O}(MNB)$, for the convenience of comparing the time complexity of regularizers, $ \mathcal{O}(MNB) $ is omitted in Table \ref{model_complexity}.}. Table \ref{model_complexity} shows that the time complexity of $ \U $-based models is much lower than that of $ \X $-based models. For example, by comparing the time complexity of the LLRT\cite{chang2017hyper} and NGmeet \cite{he2019non} methods based on the NLS prior, we can find that placing prior on $\mathcal{U}$ can reduce the time complexity. Also, by comparing LRTV \cite{he2015total}, CTV\cite{peng2022exact}, and RCTV approaches, we can also obtain the same conclusion. Additionally, by comparing NGmeet and RCTV, both of which place prior on $\mathcal{U}$ but differs in that NGmeet utilizes the NLS and L priors while our RCTV uses the LS and L priors, we can see that the time complexity of RCTV is significantly less than the NGmeet. Except for the low time complexity, our RCTV can also obtain a satisfactory denoising performance. Fig. \ref{show_introduction} illustrates the visual denoising effect of all competing methods in Table \ref{model_complexity} and one deep learning-based method on the complex noise case. From Fig. \ref{show_introduction}, it can be easily observed that our proposed RCTV achieves a relatively better denoising performance, and has the fastest execution time among the model-based methods, and even comparable with the deep learning-based approach (i.e., GRN\cite{cao2021deep}). Meanwhile, we also find that LLRT and NGmeet based on NLS prior do not attain well denoising performance on the complex noise case, which is why we choose LS prior for complex noise removal.

In summary, this paper makes the following contributions:

Firstly, we theoretically prove that when the end-member matrix $ \V $ is orthogonal, the representative coefficient matrix $ \U $ can finely inherit the spatial prior information of $ \X $. Further, the low-rank property of the $\mathbf{X}$ can be maintained by controlling the column number $R$ of $\mathbf{U}$. Therefore, the $ \U $-based regularizer can not only greatly reduce the time complexity of its corresponding HSI denoising model, but it can fully exploit HSI's spectral and spatial priors.

Secondly, based on the phenomenon that the representative coefficient matrix $ \U $ is intrinsically robust to noise perturbation and  has LS prior, we propose an RCTV regularizer. Furthermore, we use the RCTV regularization to build a model suitable for complex noise removal and then design a fast algorithm to solve this model via the FFT and the orthogonal transformation.

Thirdly, comprehensive experimental results substantiate the superiority of the proposed RCTV method beyond the state-of-the-art HSI denoising methods both on the denoising performance and execution time. Specifically, our proposed approach can obtain the best denoising results, and its running time is the shortest among all the model-based methods, and even comparable with the deep learning-based methods.

The rest of this paper is organized as follows: Section \ref{Related_work} gives some related works. In Section \ref{nota_section}, we provide some notations and preliminaries. Section \ref{ARBTV} presents the new regularization term named RCTV. Section \ref{denoising_model} introduces the proposed denoising model and its optimization algorithm. Experimental results are shown in Section \ref{Experiments_part}. Finally, conclusions are drawn in Section \ref{Conclusion_Part}.

\section{Related Work}
\label{Related_work}
In this section, we briefly review the existing model-based methods and some data-based methods.

\subsection{Model-Based HSI denoising methods}
Model-based methods are based on statistical prior modeling. Next, we will introduce the three most commonly utilized priors.

\subsubsection{HSI Denoising with Spectral Correlation Prior}
The spectral correlation prior refers to the fact that HSIs should lie in a low-dimensional spectral subspace since the images in adjacent bands are usually collected with similar sensor parameter settings. Therefore, the spectral correlation prior converts into the low-rank prior of HSI. Tools for characterizing low-rank priors can be divided into two big categories, namely matrix/tensor nuclear norm and low-rank matrix/tensor factorization. Nuclear norm as the tightest convex relaxation of matrix rank has good theoretical properties, therefore, RPCA \cite{candes2011robust} gives the exact recovery theorem for separating low-rank and sparse components from contaminated data. Since nuclear norm is convex relaxations of matrix rank, which tends not to fully characterize low rank prior of matrices. Besides, it will cost much time to solve the nuclear norm minimization problem via SVD operator when the data size is large. Therefore, many works focus on pre-specifying the rank by the low-rank matrix factorization to better characterize the low-rank prior of the matrix  \cite{cao2016robust, eriksson2010efficient, rasti2017automatic}. Although these LRMF-based models are faster, such as HyRes \cite{rasti2017automatic}, the recovery performance of these methods can be further improved as only the L prior of the data is used. In recent years, there are also some tensor-based methods to characterize L prior, such as TRPCA \cite{lu2019tensor, lu2016tensor}, CP/Tucker/Tensor-Train decomposition \cite{kolda2009tensor, sidiropoulos2017tensor, oseledets2011tensor}. The low rank tensor decomposition is often extremely time-consuming due to the large number of algebraic operators introduced in it.

\subsubsection{HSI Denoising with Non-local Similarity Prior}
The non-local spatial similarity prior refers to the fact that there always exist similar patterns in image spatial dimensions, thus this prior is often integrated into low-rank matrix/tensor approximation frameworks to remove Gaussian noise\cite{ye2014multitask, maggioni2012nonlocal, zhuang2018fast, cao2016total, xie2017kronecker, chang2017hyper}. Typical methods using such priors are KBR \cite{xie2017kronecker}, LLRT \cite{chang2017hyper}, BM4D \cite{maggioni2012nonlocal}, and NGmeet \cite{he2020non}. Since the time complexity of the SBS step is very high, which can be seen in Table \ref{prior_complexity}, and SBS is susceptible to sparse noise interference, resulting in inaccurate similar groups and inaccurate estimation of noise variance. Therefore, this kind of method is not suitable for large-scale data and mixed noise removal.
\subsubsection{HSI Denoising with Local Smoothness Prior}
The local smoothness prior is based on the fact that similar objects are often distributed in a local area with high probability. Besides, the adjacent bands of images are usually collected with similar sensor parameter settings, which results in similar values. Such local smoothness prior can be encoded by using TV regularization on the spatial and spectral domain of the HSI \cite{rudin1992nonlinear, wang2017hyperspectral}. Additionally, more advanced TV methods are proposed to characterize this prior \cite{yuan2012hyperspectral, jiang2016hyperspectral, peng2020enhanced, peng2022exact}. By integrating TV regularization into a low-rank matrix/tensor factorization framework, some classic works have been proposed and have achieved excellent results in removing mixed noise, such as LRTV \cite{he2015total}, LRTDTV \cite{wang2017hyperspectral}, LRTDGS \cite{chen2019hyperspectral}. In these models, the TV regularization is applied to each slice $ \X(:,:,i) $ for HSI data. Since the spectral band number $ B $ of HSI is too large, the time complexity of utilizing this prior will also cost much time according to analysis from Table \ref{prior_complexity}.

\subsection{HSI Denoising via Deep Learning}
Unlike model-based methods, data-based methods directly learn a denoising mapping function from a large amount of data. Utilizing the powerful image feature extraction capability of deep network, some methods were proposed to learn a non-linear end-to-end mapping between the noisy and clean HSIs. The classical deep learning methods for HSI denoising includes HSICNN\cite{yuan2018hyperspectral}, and HSIDeNet\cite{chang2018hsi}. Later, some methods based on network structure design were proposed one after another. For example, 3D-ADNet\cite{shi2021hyperspectral} applies 3D convolution instead of 2D convolution together with self-attention to extract HSI features. QRNN3D\cite{wei20203} adopts 3D convolution, quasi-recurrent pooling function to characterize the structural spatio-spectral correlation and global correlation along the spectrum dimension. \cite{lin2019hyperspectral} combines the matrix factorization method with CNN, where the CNN is a solver of the corresponding sub-problem. GRN-net\cite{cao2021deep} uses two reasoning modules to carefully extract both global and local spatial-spectral features for mixed noise removal. Further, to tackle the physical interpretability issue of the
deep neural network (DNN), SMDS-Net\cite{xiong2022smds} and T3SC\cite{bodrito2021trainable} propose a model guided spectral-spatial network by unfolding the iterative algorithm to solve the sparse model. While these networks can achieve excellent denoising performance for a given noise scale with a short test time, the test performance of these networks is dramatically affected by the training data. If the training data does not contain a certain type of noise distribution, the prediction accuracy of the learned network under this type of noise will drop significantly. In real life, it is impossible for us to know all the noise distributions in advance, and it is impossible to construct a large number of clean noise sample pairs. In a word, the generality of data-based methods is not sufficient.

To minimize the denoising time while maintaining the generalization of the model, mining prior of the representative coefficient matrix is thus a reasonable choice. Compared with mining the prior of original HSI data, the studies about mining the prior of the reprensentative coefficient is relatively rare for HSI denoising task. In fact, mining priors for representative coefficient matrices can be traced back to HSI unmixing task \cite{lillesand2015remote, yao2019nonconvex}. In recent years, some fast algorithms using priors of the representative coefficient matrix have attracted a lot of attention, such as E3DTV \cite{peng2020enhanced}, LRTFDFR \cite{zheng2020double}, NGmeet \cite{he2019non,he2020non} and many others \cite{chen2018hyperspectral, cao2019hyperspectral}.  In particular, NGmeet exploits the non-local similarity of the representative coefficient matrix to obtain the best performance on gaussian noise removal and greatly reduce the runtime for HSI denoising\footnote{The representative coefficient matrix is also called as reduced image in NGmeet\cite{he2019non}.}.

\section{Notations and Preliminaries}
\label{nota_section}
Throughout this paper, we denote scalars, vectors, matrices and tensors in light, bold lower case letters, bold upper case letters and upper cursive letters, respectively. For an $ N $-order tensor $ \mathcal{X}\in  \mathbb{R}^{I_1\times I_2\times \dots \times I_N}$, its mode-$n $ unfolding matrix is denoted as $ \X_{(n)} :=\mbox{unfold}_n(\mathcal{X}) \in \mathbb{R}^{I_n\times (I_1\cdots I_{n-1} I_{n+1}\cdots I_N)}  $, and $ \mbox{fold}_n(\X_{(n)}) = \mathcal{X}$ where $ \mbox{fold}_n $ is the inverse of unfolding operator. The mode-$ n $ product of a tensor $ \mathcal{X}\in  \mathbb{R}^{I_1\times I_2\times \dots \times I_N}$ and a matrix $ \A\in \mathbb{R}^{J_n\times I_n} $ is denoted as $ \mathcal{Y} := \mathcal{X} \times_n \A$, where $ \mathcal{Y} \in  \mathbb{R}^{I_1\times \cdots \times I_{n-1}\times J_n \times I_{n+1}\times \cdots \times I_N}$. The inner product of two matrices $ \X, \Y $ of the same size is defined as $ \left\langle \X, \Y\right\rangle := \sum_{i,j} x_{i,j}.y_{i,j}$, where $ x_{i,j} $ is $ (i,j) $-th element of $ \X $. Then the corresponding Frobenius norm and $ \ell_1 $ norm are defined as $ \| \X \|_{\scriptsize{\mbox{F}}} :=\sqrt{\left\langle \X, \X\right\rangle}$ and $ \| \X \|_1 := \sum_{i,j} |\X_{i,j}|$, respectively.

An HSI data with $B$ bands can be viewed as a three-dimensional tensor $ \mathcal{X} \in \mathbb{R}^{M\times N\times B} $. It is also represented as a Casorati matrix $ \X \in \mathbb{R}^{MN\times B} $ equipping well low-rankness (often rank $R \ll B$), whose columns comprise vectorized bands of the HSI. Besides, the tube $ \mathcal{X}(i,j,:) $ is also equal to the row $ \X(k,:) $, where $ k=(j-1)*w+i $. The total variation (TV) that effectively encodes the local smoothness prior needs to be introduced. For a gray-level image $ \X $ of size $ M\times N $, its anisotropic TV norm \cite{beck2009fast} is defined as:
\begin{equation}
\label{ani_TV}
\begin{split}
\| \X \|_{\scriptsize{\mbox{TV}}} &:= \sum_{i=1}^{M-1}\sum_{j=1}^{N-1} \{ |x_{i,j}-x_{i+1,j}| + |x_{i,j}-x_{i,j+1}|\} \\
&+\sum_{i=1}^{M-1} |x_{i,N}-x_{i+1,N}| + \sum_{j=1}^{N-1} |x_{M,j}-x_{M,j+1}|.
\end{split}
\end{equation}
It also can be written as
\begin{equation}
\label{ani_TV2}
\| \X \|_{\scriptsize{\mbox{TV}}} :=\| \mathcal{D}_h \X\|_1 + \| \mathcal{D}_v \X\|_1,
\end{equation}
where $ \mathcal{D}_h, \mathcal{D}_w $ denote the first-order forward finite-difference operator among the horizontal and  vertical direction.

\section{Representative Coefficient TV Regularization}
\label{ARBTV}
\subsection{Motivations}
For an observed noisy HSI  $ \Y \in \mathbb{R}^{MN\times B} $ contaminated by mixed noise, such as Gaussian white noise, sparse noise, stripes noise, deadlines, and missing
pixels \cite{xie2016hyperspectral, zhang2013hyperspectral, he2015hyperspectral}, it can be nearly formalized as an additive hybrid system
\begin{equation}
\Y = \X + \S + \E,
\end{equation}
where $ \X $, $ \S $ and $ \E $ are the clean HSI, sparse corruptions and other system Gaussian noise, respectively. HSI denoising task aims to estimate the clean HSI from the noisy HSI.

From the perspective of the linear spectral mixing model, each spectral signature (i.e., the row of $ \X $) can be represented by a linear combination of several end-member basis \cite{yao2019nonconvex, heylen2014review}. Considering the existing strong correlation between HSI's spectral bands, i.e., spectral low-rankness, it's reasonable to use LRMF to model the HSI as $ \X = \U \V^{\scriptsize{\mbox{T}}} $, where $  \U \in \mathbb{R}^{MN\times R} (R\ll B)$ is called the abundance map or representative coefficient basis, and $ \V \in \mathbb{R}^{B\times R}$ is called the end-member matrix or ``dictionary".  Then, the noise degradation model can be formulated as
 \begin{equation}
\Y = \U\V^{\scriptsize{\mbox{T}}} + \S + \E.
\end{equation}

Except for the low-rank prior in spectral, to further improve the denoising performance in visual, most existing works adopt some other spatial priors of the original HSI data, such as the LS prior \cite{he2015total, wang2017hyperspectral} and NLS prior \cite{chang2017hyper, he2020non}. As we claimed before, the direct prior computing on original large-sized HSI leads to the bottleneck of execution speed. As an alternative, it is stated that the spatial information of original large-sized $ \X $ can be reflected on small-sized $ \U $ to some extent through the following Theorem \ref{theorem_e}.

\begin{theorem}
\label{theorem_e}
For a rank-$R$ matrix $\X\in\mathbb{R}^{MN\times B}$ that has the factorization $ \X = \U\V^{\textrm{T}} $ with orthogonal matrix $ \V \in\mathbb{R}^{B\times R} $, its representative coefficient matrix can be obtained by $ \U=\X\V $, and we have
\begin{enumerate}
\item[a)] If the two row vectors $\X(i,:)$ and $\X(j,:)$ of original data $ \X $ are the same, then their corresponding coefficient vectors $\U(i,:)$ and $\U( j,:) $ is the same.
\item[b)] The distance and angle between any two-row vectors of $ \X $ are the same as those of the corresponding two-row vectors of $ \U $.
\end{enumerate}
\end{theorem}
The proof of Theorem \ref{theorem_e} is placed in Appendix \ref{appendix_theorem}. It demonstrates the following two things: 1) The representative coefficient of $ \X $ under $ \V $ is unique, and the same row vectors in $ \X $ must have the same representative coefficients; 2) The row spaces of the original data matrix $ \X $ and the representative coefficient matrix $ \U $ have close similarities in terms of distance, angle and equivalence property, which implies that the representative coefficient matrix will not cause substantial information loss and can well inherit the spatial information from $ \X $. Therefore, it is encouraged to explore the spatial prior on the small-sized $ \U $ instead of the big-sized $ \X $.

\begin{figure*}[!t]
		\centering
		\includegraphics[scale=0.7]{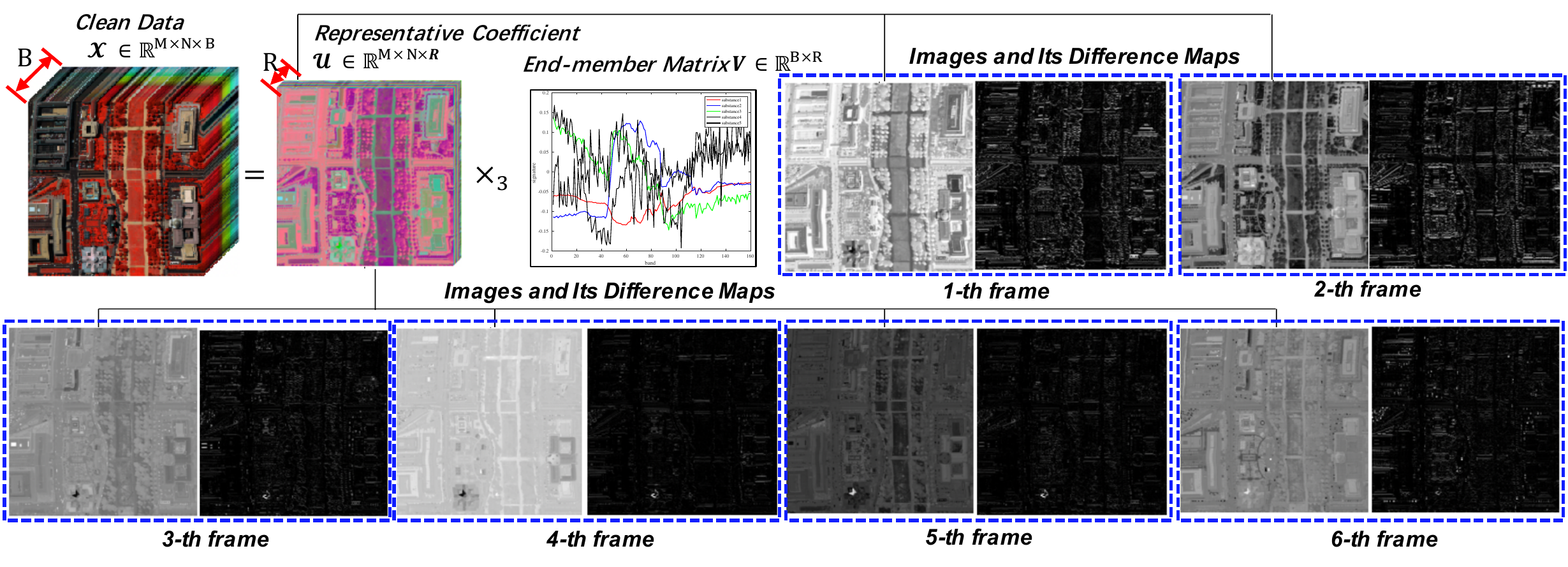}
		\vspace{-3mm}
		\caption{Illustration of the local smoothness prior for the representative coefficient matrix. Taking LRMF operator on clean data along spectral dimension, we have $ \X = \mathcal{U}\times_3\V^{\scriptsize{\mbox{T}}} $, where $ \mathcal{U}\in \mathbb{R}^{M\times N\times R}$ and $ \V \in \mathbb{R}^{B\times R}$ are the representative coefficient matrix and the end-member matrix, respectively. Here, we set $ R=6 $ and then show the images of these six slices/frames of $ \mathcal{U}$ and its difference maps. }
		\label{rbtv_illustartion}
	\end{figure*}
	
	\begin{figure*}[!]
		\centering
		\vspace{-2mm}
		\includegraphics[scale=0.48]{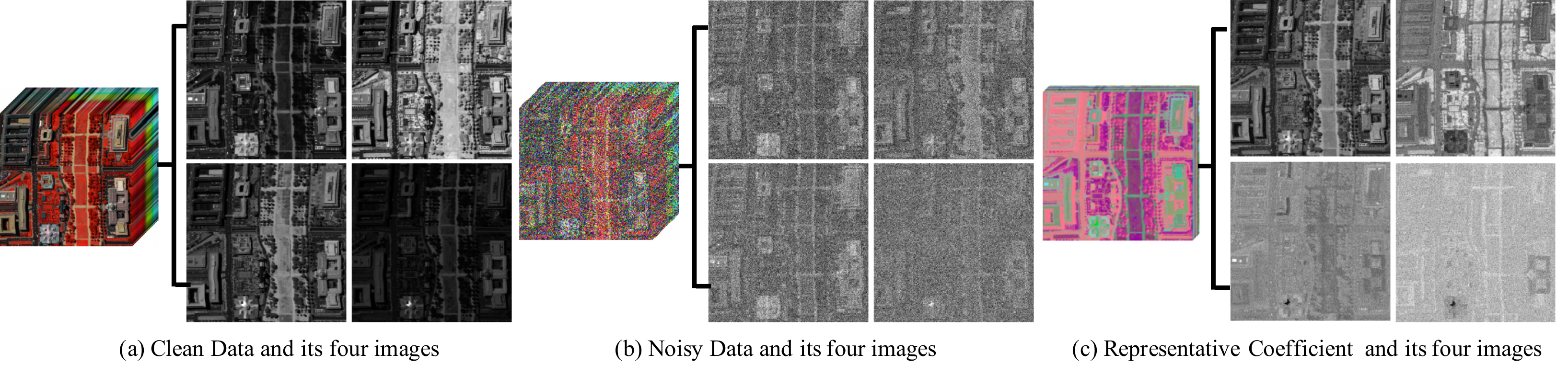}
		\vspace{-3mm}
		\caption{Here the noisy HSI is obtained by adding the Gaussian noise $\mathcal{N}(0,0.1^2)$ to each band of the clean HSI. The representative coefficient matrix is obtained by performing LRMF on the noisy HSI.}
		\vspace{-4mm}
		\label{robust_rbtv}
	\end{figure*}
	
\subsection{ RCTV Regularization }
In this subsection, we focus on modeling the LS prior of $ \U $ since such prior is common for images and meanwhile very convenient in computing.

To illustrate the existing LS priors of $ \U $, in Fig. \ref{rbtv_illustartion}, we take a real HSI data DC mall $ \mathcal{X} \in \mathbb{R}^{200\times 200\times 160} $ as an example. We first perform the LRMF operator on its unfolded matrix $ \X_{(3)} = \U \V^{\textrm{T}}$, which is equal as $  \mathcal{X} = \mathcal{U}\times_3 \V$, where $ \mathcal{U} = \mbox{fold}_3(\U)$. Then we show some spatial images of $ \mathcal{U} $ and the spectral curve in $ \V $ in Fig. \ref{rbtv_illustartion}. Since DC mall Data has a strong correlation among spectral dimension, we only need a very small number of end-members and corresponding representative coefficients to retain most of the information of the original data. As a result, the number of end-members $ R $ is set as 6. Next, we show 6 slices of the representative coefficient tensor $ \mathcal{U}$ in turn, and perform the difference operation on them. We can clearly see that the difference map of each slice is very sparse, which verifies that the representative coefficient also has strong local smoothness.

To encode the local smoothness in the slices of representative coefficient tensor $\mathcal{U} $, we impose the TV seminorm\cite{rudin1992nonlinear} on each slice of $ \mathcal{U} $ and add up them together to propose the Representative Coefficient Total Variation (RCTV) regularization. Mathematically, its definition is given below.

\begin{definition}[RCTV]
\label{definition1}
For a rank-$ R $ matrix $ \X \in \mathbb{R}^{MN \times B} $ that has the low-rank factorization $\X = \U \V^{\textrm{T}}$ with orthogonal matrix $ \V\in \mathbb{R}^{B\times R} $, we define its RCTV norm as
\begin{equation}
\| \X \|_{\scriptsize{\mbox{RCTV}}} := \sum_{i=1}^R  \| \mathcal{U}(:,:,i)\|_{\scriptsize{\mbox{TV}}}
\end{equation}
where $ \mathcal{U} = \mathcal{X} \times_3\V$. To simplify the notation, let $ \nabla_1(\U)= [\mathcal{D}_h \mathcal{U}(:,:,1),\cdots,\mathcal{D}_h \mathcal{U}(:,:,R)] $ and $ \nabla_2(\U)= [\mathcal{D}_w \mathcal{U}(:,:,1),\cdots,\mathcal{D}_w \mathcal{U}(:,:,R)] $, we have
\begin{equation}
\label{rctv_definition}
\| \X \|_{\scriptsize{\mbox{RCTV}}} :=  \| \nabla_1(\U) \|_1 + \| \nabla_2(\U) \|_1.
\end{equation}
\end{definition}

The RCTV regularization has several benefits behind. Compared with conventional TV regularization defined on original data $ \X $, the RCTV regularization defined on $ \U $ can not only maintain the low-rankness and local smoothness prior of HSI, but also greatly reduce computational complexity. Besides, there is an interesting finding that the representative coefficient $ \U $ is robust to noise disturbance, indicating the RCTV has huge potential in HSI denoising tasks.

The robustness of the representative coefficient matrix $ \U $ comes from the low-rank decomposition. The effects of low-rank matrix factorization and singular value truncation are similar. PCA/SVD can remove noise to a certain extent and obtain principal components \cite{abdi2010principal, wold1987principal, wall2003singular}. Therefore, $ \U $ has been denoised to some extent compared to the original noisy data, and the noise contained in $ \U $ is lighter than the noisy data $ \Y $. To better illustrate this, we selected four images of clean data, noisy data, and representative coefficient matrix for display in Fig. \ref{robust_rbtv}. Compared with the original images (i.e., Fig. \ref{robust_rbtv}(a)), we can see that the structure in the noisy images (i.e., Fig. \ref{robust_rbtv}(b)) have been severely damaged due to the interference of noise, but the images of representative coefficient (i.e., Fig. \ref{robust_rbtv}(c)) can still clearly retain the structural information of the original data. Such robustness of the representative coefficient matrix is also analyzed in E3DTV regularization \cite{peng2020enhanced}.

\section{HSI mixed denosing via RCTV regularization}
\label{denoising_model}
\subsection{Proposed Model}
In this section, we propose our denoising model based on modeling the noise. The noise structures in practical HSIs are very complex. There are some works about modeling the noise distributions, such as a mixture of Gaussian (MoG) \cite{meng2013robust, chen2017denoising}, and mixture of Laplacian \cite{cao2015low, cao2016robust}. Considering that the noise structures in practical HSIs can be roughly divided into two types. One type is sparse noise, such as stripes noise, deadlines, and missing pixels\cite{wang2017hyperspectral}, and the other type is Gaussian noise. Here, we simply use $ \ell_1 $-norm and $ \ell_2 $-norm to characterize the sparse and Gaussian noise, respectively, which is similar to the modeling mechanism in LRTDTV \cite{wang2017hyperspectral}, and LRTFDFR \cite{zheng2020double}. According to the definition of RCTV with Eq. (\ref{rctv_definition}), we propose the denoising model based on RCTV regularization as follows:
\vspace{-1mm}
\begin{equation}
\label{denoise_model}
\begin{split}
\min_{\U,\V,\E,\S} \ & \sum_{i=1}^2 \tau_i \| \nabla_i(\U) \|_1 + \beta \| \E\|_{\scriptsize{\mbox{F}}}^2 + \lambda \| \S\|_1, \\
s.t. \ & \Y = \U\V^{\scriptsize{\mbox{T}}} + \E + \S,\V^T\V=\mathbf{I},
\end{split}
\end{equation}
where $ \tau_i,  \beta , \lambda$ are trade-off parameters to balance the weight  of each term in the objective.

It is worth noting that the proposed RCTV regularization is robust to noise to some extent and can fully capture the spatial information of representative coefficient matrix $ \U $. Thus this model is expected to have a strong ability for mixed noise removal. Since the size of $ \U $ is much lower than the original HSI data $ \X $, the time complexity of this model is extremely low, so we can expect this model to bring robust and fast denoising results for mixed noise removal.

\subsection{Optimization Algorithm}
By introducing the auxiliary variables $ \G_1,\G_2$, we rewrite model (\ref{denoise_model}) as the following equivalent formulation:
\vspace{-2mm}
\begin{equation}
\label{alm_model}
\begin{split}
\min_{\U,\V,\E,\S} \ &  \sum_{i=1}^2 \tau_i\|\G_i\|_1 + \beta \| \E\|_{\scriptsize{\mbox{F}}}^2 + \lambda \| \S\|_1, \\
s.t. \ & \nabla_i(\U) = \G_i, i=1,2, \\
 \ & \Y = \U\V^{\scriptsize{\mbox{T}}} + \E + \S,\V^T\V=\mathbf{I}.
\end{split}
\end{equation}
\vspace{-4mm}

Then we use the well-known alternating direction method
of multipliers (ADMM) to derive an efficient algorithm for solving model (\ref{alm_model}). The augmented Lagrangian function of model (\ref{alm_model}) is:
\begin{equation}
\label{alm_reb}
\begin{split}
& \mathcal{L}(\U,\V,\E,\S,\{\G\}_{i=1}^2,\{\mathbf{\Gamma}_i\}_{i=1}^3) :=\sum_{i=1}^2 \tau_i\| \G_i \|_1 \\
 & \quad +\frac{\mu}{2} \sum_{i=1}^2\| \nabla_i(\U)- \G_i + \frac{\mathbf{\Gamma}_i}{\mu}\|_F^2 + \beta \| \E \|_{\scriptsize{\mbox{F}}}^2 + \lambda \| \S \|_1 \\
& \quad  +  \frac{\mu}{2} \| \Y - \U\V^{\scriptsize{\mbox{T}}} -\E -\S + \frac{\mathbf{\Gamma}_3}{\mu}\|_F^2, \\
\end{split}
\end{equation}
where $ \mu $ is the penalty parameter, $ \{\mathbf{\Gamma_i\}}_{i=1}^3 $ are the Lagrange multipliers. We then discuss how to solve its sub-problems for each involved variable.

\textbf{Update} $\G_i,(i=1,2)$. Fixing other variables except $\G_i$ in Eq. (\ref{alm_reb}), we obtain the following sub-problem:
\begin{equation}
\arg \min_{\G_i} \tau_i \| \G_i \|_1 + \frac{\mu}{2} \| \nabla_i (\U) - \G_i + \frac{\mathbf{\Gamma}_i}{\mu}\|_F^2.\\
\end{equation}
Utilizing the soft threshold operator $ \mathcal{S} $ defined by \cite{donoho1995noising}, we have
\begin{equation}
\label{slover_G}
\G_i = \mathcal{S}_{\tau_i/\mu}\left( \nabla_i(\U) + \mathbf{\Gamma_i}/\mu \right).\\
\end{equation}

\textbf{Update} $\V$. Fixing other variables except $ \V $, we have
\begin{equation}
\label{solveV}
\max_{\V^{\textrm{T}}\V=\mathbf{I}} \left\langle (\Y-\E-\S+\mathbf{\Gamma}_3/\mu)^{\scriptsize{\mbox{T}}}\U,\V \right\rangle.
\end{equation}
According to the Theorem 1 in \cite{peng2020enhanced}, we can get the solution of Eq. (\ref{solveV}) as follows:
\begin{equation}
\label{slover_V}
\left\{
\begin{split}
& \left[\mathbf{B},\mathbf{D},\mathbf{C}\right]= \mbox{svd}(  (\Y-\E-\S+\mathbf{\Gamma}_3/\mu)^{\scriptsize{\mbox{T}}}\U ),\\
& \V=\mathbf{B}\mathbf{C}^{\textrm{T}}.
\end{split}
\right.
\end{equation}

\textbf{Update} $\U$. Deriving Eq. (\ref{alm_reb}) with respect to  $\U$, we can get the following linear system:
\begin{equation}
\label{nabla_update}
\begin{split}
\left(\mu\mathbf{I}+\mu\sum_{i=1}^2\mathbf{\nabla}_i^{\scriptsize{\mbox{T}}}\mathbf{\nabla}_i\right)(\U) & = \mu(\Y-\E-\S+\mathbf{\Gamma}_3/\mu)\V\\
& +\sum_{i=1}^2\mathbf{\nabla}_i^{\scriptsize{\mbox{T}}}\left(\mu \G_i-\mathbf{\Gamma}_i\right).\\
\end{split}
\end{equation}
where  $\mathbf{\nabla}_i^{\textrm{T}}(\cdot)$ indicates the `transposition' operator of ${\nabla}_i(\cdot)$\footnote{Since $ \mathbf{\nabla}_i(\cdot)$ is a linear operator on $\U$, there exists a matrix $\A_i$ which makes the operation $\A_i\cdot\mbox{vec}(\U)$ equivalent to $ \mathbf{\nabla}_i(\U)$. Then,  $ \mathbf{\nabla}_i^{\textrm{T}}(\cdot)$ means the operator equivalent to the transposed matrix $\A_i^{\textrm{T}}$.}. Taking the difference operation  $\mathbf{\nabla}_i(\U)$ as the convolution to a difference filter $\mathbf{D}_i\otimes\mbox{fold}(\U)$, where $\mathbf{D}_i$ is the correlated difference filter\footnote{For example, $[1,-1]$ and $[1;-1]$ for the horizontal and vertical difference filters, respectively.}, we can readily employ fast Fourier transform to efficiently solve Eq. (\ref{nabla_update}). Specifically, by performing Fourier transform on both sides of the equation and adopting the convolution theorem, the closed-form solution to $\U$ can be easily deduced as \cite{krishnan2009fast} :
\begin{equation}
\label{slover_U}
\left\{
\begin{split}
&{\mathbf{H}}=\sum_{i=1}^2\mathcal{F}\left(\mathbf{D}_i\right)^*\!\odot\!\mathcal{F}\left(\mbox{fold}\left(\mu\G_i)- \mathbf{\Gamma}_i\right)\right),\\
&\mathbf{T}_x=|\mathcal{F}(\mathbf{D}_1)|^2+\mathcal{F}(\mathbf{D}_2)|^2, \\
&\mathbf{X}=\mathcal{F}^{-1}\left(\frac{ \mathcal{F}\left(\mbox{fold}(\mu(\Y-\E-\S)+\mathbf{\Gamma}_3) \right) +\mathbf{H}}{\mu{\mathbf{1}}+\mu\mathbf{T}_x}\right),
\end{split}
\right.
\end{equation}
where $\mathbf{1}$ represents the tensor with all  elements as $1$, $\odot$ is the element-wise multiplication, $\mathcal{F}(\cdot)$ is the Fourier transform, and
$\left|\cdot\right|^2$ is the element-wise square operation.

\textbf{Update} $\E$. Extracting all items containing $\E$ in Eq. (\ref{alm_reb}), we can get:
\begin{equation}
\label{slover_E}
\E = (\mu(\Y-\U\V^{\textrm{T}}-\S) +\mathbf{\Gamma}_3)/(\mu+2\beta).
\end{equation}

\textbf{Update} $\S$. Extracting all items containing $\S$ in Eq. (\ref{alm_reb}), we can get:
\begin{equation}
\label{slover_S}
\S = \mathcal{S}_{\lambda/\mu} (\Y-\U\V^{\textrm{T}}-\E+\mathbf{\Gamma}_3/\mu).
\end{equation}
	
\textbf{Update} multipliers $ \mathbf{\Gamma}_i,(i=1,2,3)$.
Based on the general ADMM principle, the multipliers are
further updated by the following equations:
\begin{equation}
\label{slover_M}
\left\{
\begin{split}
\mathbf{\Gamma}_i &=\mathbf{\Gamma}_i+\mu\left(\nabla_i(\U)-\G_i\right),i=1,2,\\
\mathbf{\Gamma}_3&=\mathbf{\Gamma}_3+\mu\left(\Y -\U\V^{\textrm{T}} - \E -\S \right), \\
\mu & = \mu\rho,\ \\
\end{split}
\right.
\end{equation}
where $ \rho $ is a constant value greater than 1.

Summarizing the aforementioned descriptions, we can get the following {Algorithm} \ref{alg_denoise}.

\begin{algorithm}
\caption{RCTV Solver.}\label{alg_denoise}
\small
\begin{algorithmic}[1]
\renewcommand{\algorithmicrequire}{\textbf{Input:}}
\renewcommand{\algorithmicensure}{\textbf{End}}
\REQUIRE HSI data $\Y\in\mathbb{R}^{MN\times B}$, the hyper-parameter $ \tau=0.01 $, $ \lambda $ and rank $ R $, the parameters in ADMM framework are set as: $\mu=1e-3,\rho=1.25 $, and $ \epsilon = 10^{-6}$.
\renewcommand{\algorithmicrequire}{\textbf{Initialization:}}
\renewcommand{\algorithmicensure}{\textbf{End}}
\REQUIRE Initial $\U = \U_R\mathbf{\Sigma}_R, \V=\V_R$, where $ \U\mathbf{\Sigma}\V$ is the SVD operation of $\Y$ and $ \U_R,  \mathbf{\Sigma}_R, \V_R$ are the first $ R $ vectors.
\WHILE {not converge}
\STATE
Update the variables $\G_i$, $\V $,  $\U$, $ \E $ and $\S$ by Eq. (\ref{slover_G}), (\ref{slover_V}), (\ref{slover_U}), (\ref{slover_E}) and (\ref{slover_S}).
\STATE
Update $\mathbf{\Gamma}_i,(i=1,2,3)$ by Eq. (\ref{slover_M}), and $ \mu = \rho \mu $.
\STATE
Check the convergence conditions\\
$\quad \| \Y-\U\V^{\textrm{T}}-\E -\S \|_{\textrm{F}}^2/\|\Y\|_{\textrm{F}}^2\leq\epsilon$,\\
$\quad \Vert\nabla_{i}\U-\G_i\|_{\textrm{F}}^2/\| \Y \|_{\textrm{F}}^2\leq\epsilon,i=1,2$.\\
\ENDWHILE
\renewcommand{\algorithmicrequire}{\textbf{Output:}}
\renewcommand{\algorithmicensure}{\textbf{End}}
\REQUIRE  Recovered Data $\mbox{fold}_3(\U\V^{\textrm{T}})\in\mathbb{R}^{M\times N\times B}$.
\end{algorithmic}
\end{algorithm}

\subsection{Computational Complexity Analysis}
As shown in Algorithm \ref{alg_denoise}, the computational cost of each iteration consists of updating $\U$ via cheap FFT, updating $\V$ via small-scale SVD computation, updating $\E$ and $\S$ via soft threshold operation, and some matrix multiplications. We have known that the time complexity of the soft threshold operator is $ \mathcal{O}(MNB) $, the time complexity of solving $ \V $ is $ \mathcal{O}(BR^2) $, and the time complexity of solving $ \U $ is $ \mathcal{O}(MNR\log(MN)) $ from Table \ref{prior_complexity}, so the time complexity of Algorithm \ref{alg_denoise} is $ \mathcal{O}(MNB+BR^2+RMN\log(MN)) $, which is also given in Table \ref{model_complexity}. 

Different from modeling on the original HSI data $ \X $, we can avoid large-scale SVD calculation and only need to solve $ R $ TV sub-problems by adding TV regularization to $\mathbf{U}$. Since $R/B$ is quite small (usually between 0.05$\sim$0.15), the time complexity of Algorithm \ref{alg_denoise} based on RCTV regularization thus becomes extremely low.

\section{Experiments}	
\label{Experiments_part}	
To demonstrate the effectiveness of our proposed algorithm, we compare our denoising results with 14 state-of-art denoising methods on both synthetic and real HSI data. These methods can be divided into four categories. Specifically, for models based on pure L prior, we choose WNNM-RPCA  \cite{gu2017weighted}\footnote{The WNNM model was first proposed in \cite{gu2014weighted}. Here we use the RPCA version of WNNM, specifically, replacing the nuclear norm in RPCA with the weight nuclear norm.} based on weighted nuclear norm minimization, LRMR \cite{zhang2013hyperspectral} based on patch denoising, and TDL \cite{peng2014decomposable} based on dictionary learning. For models based on NLS prior, we choose KBR \cite{xie2017kronecker}, LLRT \cite{chang2017hyper} based on modeling nonlocal similarity on original HSI data, NGmeet \cite{he2020non} based on modeling nonlocal similarity on coefficient matrix. For models based on joint L and LS priors, we choose some robust denoising methods for complex noise removal, such as LRTV \cite {he2015total} and LRTDTV \cite{wang2017hyperspectral}, E3DTV \cite{peng2020enhanced}, CTV-RPCA \cite{peng2022exact}, LRTFDFR \cite{zheng2020double}. For the models based on deep learning (DL), we choose the aforementioned three deep learning methods, i.e., HSI-DeNet \cite {chang2018hsi}, HSICNN \cite{yuan2018hyperspectral}, and GRN-net \cite{cao2021deep}. All relevant parameters in these competing methods are fine-tuned by their given default settings or by following the rules in corresponding papers to guarantee their possibly optimal performance. Four quantitative quality indices are employed: MPSNR, MSSIM, ERGAS and MSAM, where MPSNR and MSSIM are the mean value of PSNR and SSIM on each band. PSNR and SSIM are two conventional spatial-based metrics for a single image, while ERGAS and MSAM are spectral-based evaluation measures. The larger MPSNR and MSSIM are, and the smaller ERGAS and MSAM are, the better quality the restored images become. Before the experiment, the gray values of the data are re-scaled to [0,1] via max-min formula band by band. All the experiments are conducted on a PC with Intel Core i5-10600KF CPU@4.10GHz, NVIDIA RTX 3080 GPU, and 32-GB memory.

%\begin{table}[htbp]
%\renewcommand{\arraystretch}{1.15}
%\setlength\tabcolsep{3.0pt}
%\footnotesize
%  \caption{Categories of related methods for HSI denoising. }
%  \label{compare_model}
%  \setlength{\abovecaptionskip}{5pt}
%  \setlength{\belowcaptionskip}{5pt}
%  \centering
%  \vspace{-0.2cm}
%\begin{tabular}{l|c}
%\Xhline{1.2pt}
%Types & Methods \\
%\hline
%\hline
%\textbf{L} & WNNM \cite{gu2014weighted, gu2017weighted}, LRMR \cite{zhang2013hyperspectral}, HyRes \cite{rasti2017automatic} \\
%\hline
%\textbf{L\& NNS} &  BM4D\cite{maggioni2012nonlocal}, TDL \cite{peng2014decomposable}, LLRT \cite{chang2017hyper}, Ngmeet \cite{he2020non}\\
%\hline
%\multirow{2}{*}{\makecell[c]{\textbf{L\& LS}}} &
%LRTV \cite {he2015total}, LRTDTV \cite{wang2017hyperspectral}, E3DTV \cite{peng2020enhanced}, \\
% & CTV-RPCA \cite{peng2022exact}, LRTD-DFR \cite{zheng2020double}\\
%\hline
%\textbf{DL} & HSI-DeNet \cite {chang2018hsi}, HSICNN \cite{yuan2018hyperspectral}, GRN-net \cite{cao2021deep}\\
%\Xhline{1.2pt}
%\end{tabular}
%\end{table}

 \begin{figure}[!]
		\centering
		\includegraphics[scale=0.48]{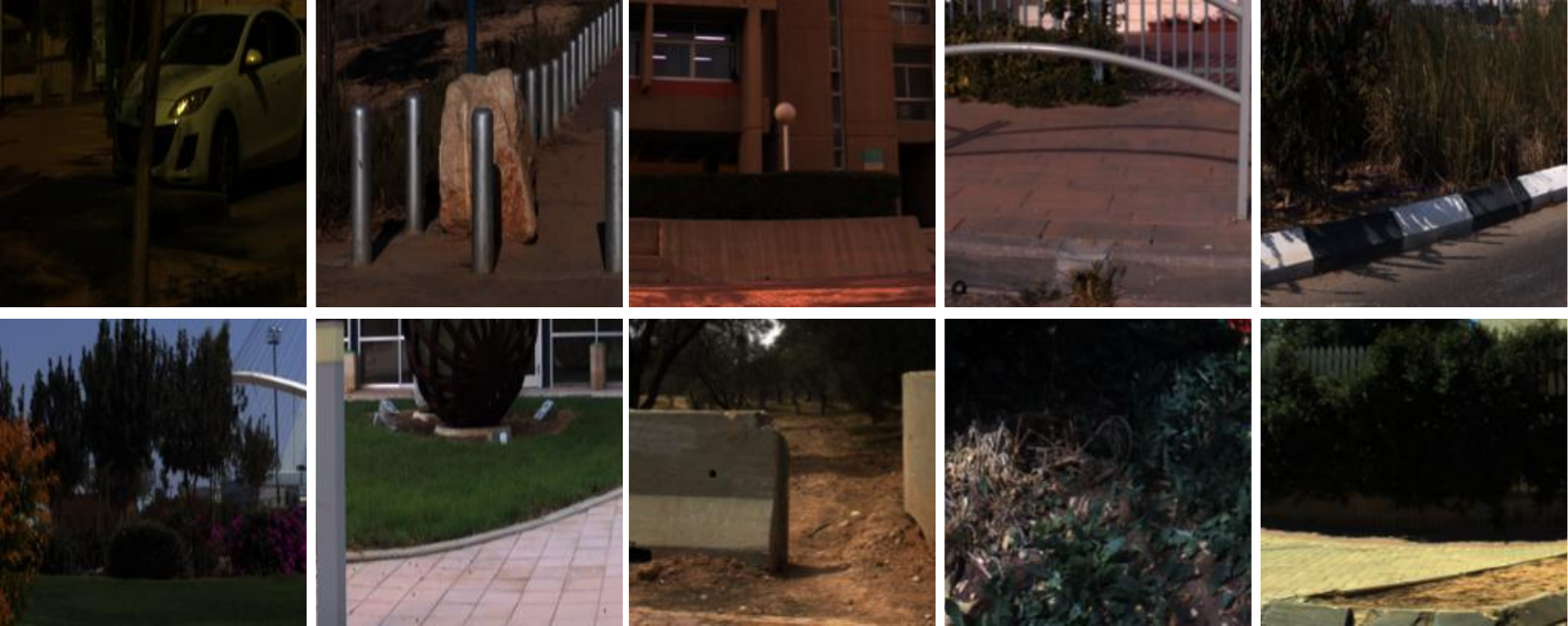}
		\vspace{-2mm}
		\caption{Ten test images in ICVL dataset.}
		\label{icvl_test}
	\end{figure}
	
	 \begin{figure}[!]
		\centering
		\includegraphics[scale=0.48]{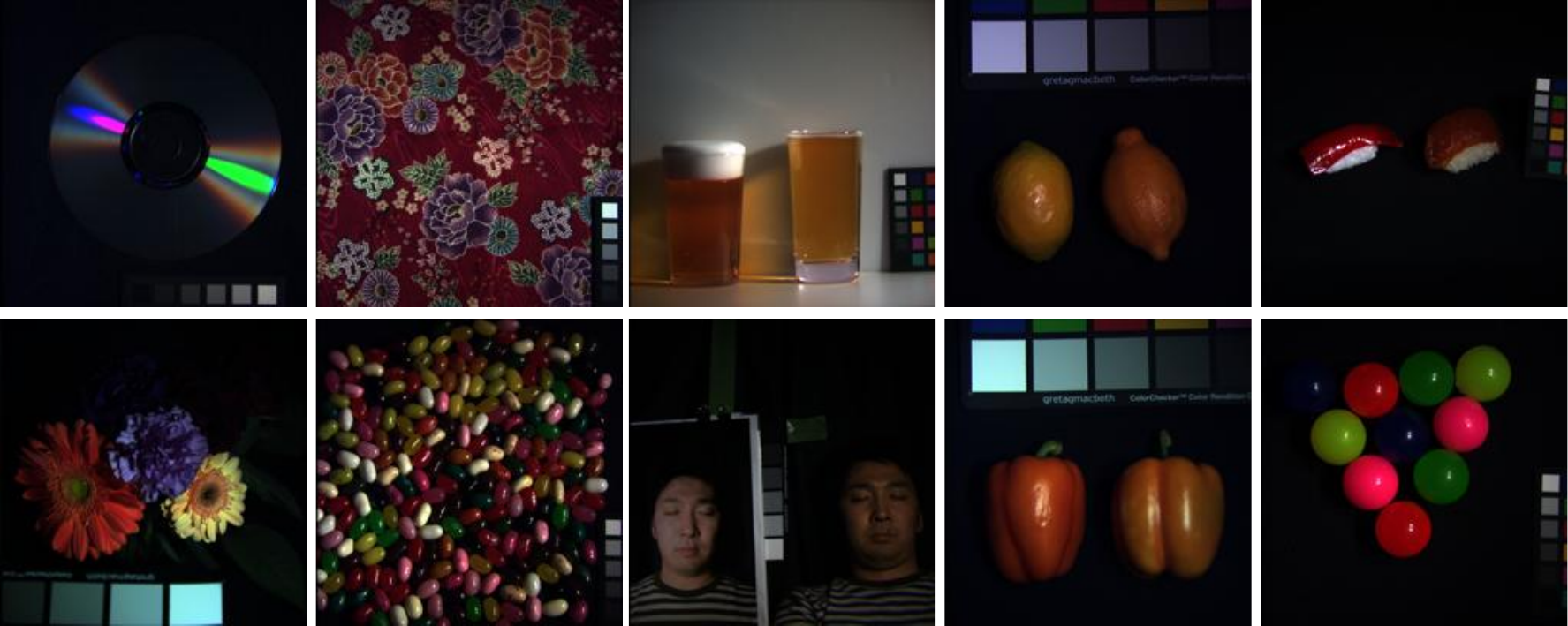}
		\vspace{-2mm}
		\caption{Ten test images in CAVE dataset.}
		\label{cave_test}
	\end{figure}

\subsection{Synthetic Experiments}
This experiment aims to evaluate the performance of the
proposed model based on RCTV regularization with other competing methods quantitatively. For a full comparison, in this section, we select two clean multispectral image (MSI) datasets, i.e., ICVL dataset \footnote{It is available at \url{http://icvl.cs.bgu.ac.il/hyperspectral/}}, and CAVE dataset \footnote{\url{https://www.cs.columbia.edu/CAVE/databases/multispectral/}}, and one HSI data DC mall for comparative experiments.

Specifically, the ICVL dataset contains 201 images, each of which is with a size of $ 1392 \times 1300 \times 31$. Since we choose some deep learning models for comparison, to be fair, we need to retrain these models. Here, we select 30 images with a size of $ 512\times 512 \times 31 $ as training data and train each model according to the original network settings.  We choose ten images with a size of $ 512\times 512 \times 31 $ as test data to evaluate all methods, which are shown in Fig. \ref{icvl_test}. The CAVE dataset contains 32 images, each of which is with a size of $ 512 \times 512 \times 31 $.  We also choose ten images as test data to evaluate all methods, which are shown in Fig. \ref{cave_test}. The pure DC mall is a common HSI data with a size of $ 200\times 200\times 160 $, which is used in many works\cite{zhang2013hyperspectral, peng2020enhanced }.

Since real HSIs are always polluted by complex noise, such as Gaussian noise, sparse noise (stripe, deadline, or impulse noise), or a mixture of them, thus we add six kinds of complex noise into the clean images to simulate these real noise cases, which are shown as follows:
\begin{enumerate}
\item[ \textbf{ (a)}] The i.i.d Gaussian noises with standard variance $ \sigma =0.1 $ are added to each band of MSI data in the ICVL, the CAVE dataset and the DC mall data.
\item[\textbf{ (b)}] First add the same Gaussian noise as in case (a), then the deadlines are added to some bands. For the ICVL and CAVE datasets, the deadlines are added from band 11 to 20, with the number of stripes randomly selected from 5 to 55, and the width of each stripe is randomly generated from 1 to 5. For the DC mall Data, the deadlines are added from band 91 to band 130, with the number of stripes randomly selected from 3 to 10, and the width of each stripe is randomly generated from 1 to 3.
\item[\textbf{ (c)}] A mixture of i.i.d Gaussian noise with standard variance $ \sigma =0.075 $ and salt and pepper noise with a proportion of $ s=0.1 $ are added to each band of the MSI data in ICVL, CAVE dataset, and DC mall data.
\item[ \textbf{ (d)}] First add the same noise as in case (c), then add additional deadlines as in case (b).
\item[ \textbf{ (e)}] A mixture of Gaussian and salt and pepper noise are added to different bands with different degrees. Here, the standard variance of Gaussian noise and percentage of salt and pepper noise are randomly selected from the interval [0.05,0.15]. Additionally, we also add deadlines as case (b) does.
\item[ \textbf{ (f)}]  First add the same noise as in case (e), then add stripe noise to some bands. Specifically, for the MSI data in ICVL/CAVE dataset, the stripes are added from band 21 to band 30, with the number of stripes being randomly selected from 50 to 100. For DC mall, the stripes are added from band 141 to band 160, with the number of stripes being randomly selected from 20 to 40.
\end{enumerate}

\begin{table*}[htbp]
\renewcommand{\arraystretch}{1.15}
\setlength\tabcolsep{3.0pt}
\footnotesize
  \caption{Quantitative comparison of all competing methods under different levels of noises in ten testing \textbf{ICVL} images. The best and second results are highlighted in bold and \underline{underline}, respectively.}
  \label{icvl_table}
  \setlength{\abovecaptionskip}{5pt}
  \setlength{\belowcaptionskip}{5pt}
  \centering
  \vspace{-0.2cm}
\begin{tabular}{l|c|c||c|c|c||c|c|c||c|c|c||c|c|c|c|c|c}
     \Xhline{1pt}
     	\multirow{3}{*}{\makecell[c]{Noise\\Types}} &\multirow{3}{*}{Metric}& \multirow{3}{*}{Noisy}& \multicolumn{3}{c||}{ \textbf{L}} & \multicolumn{3}{c||}{ \textbf{L \& NLS}} & \multicolumn{3}{c||}{ \textbf{DL}}  & \multicolumn{6}{c}{ \textbf{L\& LS}}  \\
     	\cline{4-18}
     	& & &WNNM  &\multirow{2}{*}{LRMR} & \multirow{2}{*}{TDL} &\multirow{2}{*}{KBR}&\multirow{2}{*}{LLRT}& NG-  &HSI- &HSI- &GRN & \multirow{2}{*}{LRTV}  &LRTD &\multirow{2}{*}{E3DTV}  & CTV- & LRTF-& \multirow{2}{*}{\textbf{RCTV}}\\
		& & &-RPCA & & & & & Meet & CNN & DeNet & -Net & & -TV & & RPCA &DFR \\
		\Xhline{1pt}
\multirow{4}{*}{\makecell[c]{Case\\(a)}}&MSPNR &20.00 & 31.84& 32.56 & 38.81& 40.62 & 40.89& \textbf{41.75}& 41.12& 40.91& \underline{41.28}& 35.43& 36.61& 37.42& 35.42& 37.62 &  38.37 \\
&MSSIM& 0.465& 0.926 & 0.895 & 0.968 &  0.975 &  0.976 & \textbf{0.980} & 0.977 & 0.975 & \underline{0.979} & 0.935 & 0.956 & 0.961 & 0.942 & 0.962& 0.963 \\
&ERGAS& 410.1& 115.2 & 128.7 & 44.92 & 39.42 & 38.56 & \textbf{31.86} & 35.68 & 36.63 & \underline{33.36} & 72.15 & 62.75 & 63.85 & 74.19 & 48.28& 45.99 \\
&MSAM& 0.693 & 0.270& 0.299& 0.082& 0.068& \textbf{0.061}& \textbf{0.056}& 0.065& 0.067& 0.062& 0.473& 0.217& 0.112& 0.122& 0.114& 0.075\\
\Xhline{1\arrayrulewidth}
\multirow{4}{*}{\makecell[c]{Case\\(b)}}&MSPNR& 19.73&31.77& 32.32 & 37.71& 39.76& 39.87& \textbf{40.79}& 40.37& 39.91& \underline{40.57}& 34.45& 35.69& 36.43& 35.26& 37.17& 37.77\\
&MSSIM &0.454 & 0.926& 0.894& 0.962& 0.968 & 0.970& \textbf{0.977}& 0.968& 0.955& \underline{0.975}& 0.924& 0.950& 0.955& 0.942& 0.960& 0.961 \\
&ERGAS& 422.3 & 115.7& 131.0& 156.9& 747.9 & 45.9& \underline{39.7}& 46.5& 42.6& \textbf{36.6}& 85.7& 69.9& 69.4& 75.2& 50.6& 49.7 \\
&MSAM& 0.704& 0.225& 0.263&  0.089& 0.077 & 0.076& \textbf{0.075}& 0.082& 0.079& \underline{0.077}& 0.439& 0.197& 0.103& 0.105& 0.105& 0.081 \\
\Xhline{1\arrayrulewidth}
\multirow{4}{*}{\makecell[c]{Case\\(c)}}&MSPNR& 12.16 & 31.04& 31.41& 23.95& 21.85& 22.97& 23.01& 35.12& 34.96& 37.05& 34.84& 36.80& \underline{37.63}& 35.84& 36.56& \textbf{37.87}\\
&MSSIM& 0.162& 0.902& 0.873& 0.720 &  0.586 & 0.632& 0.655& 0.896& 0.892& 0.948& 0.940& 0.951& \textbf{0.962}& 0.947& 0.950& \underline{0.961}\\
&ERGAS& 1029 & 114.3& 111.3&  281.4& 420.3 & 328.6& 306.7& 128.5& 132.4& 62.6& 143.9& 61.2& \underline{59.2}& 69.4& 65.6& \textbf{52.3} \\
&MSAM& 0.815 & 0.194& 0.158& 0.164& 0.263& 0.282& 0.267& 0.111& 0.112& 0.097& 0.191& 0.115& \underline{0.119}& 0.134& 0.155& \textbf{0.083}\\
\Xhline{1\arrayrulewidth}
\multirow{4}{*}{\makecell[c]{Case\\(d)}}&MSPNR& 12.12 & 30.82& 31.11&  23.25& 21.55 &22.24& 22.47& 34.82& 34.59& 36.75& 34.64& 36.50& \underline{37.33}& 35.58& 36.12& \textbf{37.49}\\
&MSSIM& 0.158 & 0.901& 0.872& 0.718& 0.585& 0.630& 0.641& 0.893& 0.890& 0.946& 0.939& 0.949& \textbf{0.960}& 0.945& 0.948& \underline{0.959}\\
&ERGAS& 1031 & 118.7& 115.6& 297.3&  440.2 & 341.3& 395.2& 134.2& 138.7& 65.1& 148.6& 63.8& \underline{61.8}& 72.4& 69.7& \textbf{55.5} \\
&MSAM& 0.822 & 0.211& 0.176& 0.188& 0.282 &0.304& 0.291& 0.127& 0.129& 0.112& 0.216& 0.142& \underline{0.132}& 0.152& 0.182& \textbf{0.107}\\
\Xhline{1\arrayrulewidth}
\multirow{4}{*}{\makecell[c]{Case\\(e)}}&MSPNR& 13.56 & 30.42& 30.34 & 24.93 & 22.25 & 24.48& 24.76& 34.53& 34.37& 36.62& 34.51& 36.28& \underline{36.59}& 35.52& 35.88& \textbf{36.93} \\
&MSSIM& 0.201 & 0.881& 0.838& 0.704& 0.594 & 0.642& 0.657& 0.904& 0.901& 0.938& 0.934& 0.939& \textbf{0.948}& 0.932& 0.935& \underline{0.946} \\
&ERGAS& 883.1 & 129.6& 144.3&  287.0& 334.5 & 319.2& 297.0& 135.2& 138.9& 74.2& 138.3& 70.7& \underline{69.1}& 76.8& 73.4& \textbf{66.9} \\
&MSAM& 0.783 & 0.247& 0.202&  0.236& 0.289 & 0.276& 0.261& 0.147& 0.142& \textbf{0.124}& 0.199& 0.131& 0.140& 0.157& 0.161& \underline{0.129}\\
\Xhline{1\arrayrulewidth}
\multirow{4}{*}{\makecell[c]{Case\\(f)}}&MSPNR&13.15& 30.01& 30.12 & 24.90 & 22.15 & 24.19& 24.33& 34.29& 34.31& \underline{36.52} & 33.32& 36.03& 36.44& 35.14& 35.55& \textbf{36.76} \\
&MSSIM&0.197 & 0.869& 0.831& 0.686& 0.617& 0.648& 0.653& 0.902& 0.901& 0.935& 0.911& 0.934& \underline{0.944}& 0.926& 0.929& \textbf{0.946}\\
&ERGAS&888.2& 135.7& 141.6&  283.7& 344.6 & 331.4& 318.6& 138.9& 137.6& 76.8& 142.5& 73.3& \underline{70.0}& 78.9& 72.5& \textbf{67.3} \\
&MSAM&0.788& 0.256& 0.212& 0.248& 0.313 & 0.295& 0.289& 0.163& 0.168& \textbf{0.135}& 0.218& 0.141& 0.152& 0.171& 0.176& \underline{0.138}\\
\Xhline{1\arrayrulewidth}
\hline
\hline
\Xhline{1\arrayrulewidth}
\multicolumn{2}{c|}{ Mean Times(s)}& & 14.1& 42.9&  59.8& 1632.5 & 1547.7& 207.3& 11.4& \underline{11.2}& \textbf{10.98}& 251.6& 195.3& 63.5& 116.2& 130.3& 11.7\\
\Xhline{1\arrayrulewidth}
          \Xhline{1pt}
\end{tabular}
\vspace{-0.2cm}
\end{table*}

\begin{table*}[htbp]
\renewcommand{\arraystretch}{1.15}
\setlength\tabcolsep{3.0pt}
\footnotesize
  \caption{Quantitative comparison of all competing methods under different levels of noises in ten testing \textbf{CAVE} images. The best and second results are highlighted in bold and \underline{underline}, respectively.}
  \label{cave_table}
  \setlength{\abovecaptionskip}{5pt}
  \setlength{\belowcaptionskip}{5pt}
  \centering
  \vspace{-0.2cm}
\begin{tabular}{l|c|c||c|c|c||c|c|c||c|c|c||c|c|c|c|c|c}
     \Xhline{1pt}
     	\multirow{3}{*}{\makecell[c]{Noise\\Types}} &\multirow{3}{*}{Metric}& \multirow{3}{*}{Noisy}& \multicolumn{3}{c||}{ \textbf{L}} & \multicolumn{3}{c||}{ \textbf{L \& NLS}} & \multicolumn{3}{c||}{ \textbf{DL}}  & \multicolumn{6}{c}{ \textbf{L\& LS}}  \\
     	\cline{4-18}
     	& & &WNNM  &\multirow{2}{*}{LRMR} & \multirow{2}{*}{TDL} &\multirow{2}{*}{KBR}&\multirow{2}{*}{LLRT}& NG-  &HSI- &HSI- &GRN & \multirow{2}{*}{LRTV}  &LRTD &\multirow{2}{*}{E3DTV}  & CTV- & LRTF-& \multirow{2}{*}{\textbf{RCTV}}\\
		& & &-RPCA & & & & & Meet & CNN & DeNet & -Net & & -TV & & RPCA &DFR \\
		\Xhline{1pt}
\multirow{4}{*}{\makecell[c]{Case\\(a)}}&MSPNR&20.00& 29.28& 31.43 & 38.92& 40.92 & \textbf{41.89}& \underline{41.49}& 40.76& 40.89& 41.15& 33.31& 36.10& 37.47& 33.89& 35.72& 37.70\\
&MSSIM&0.417 & 0.849& 0.867 & 0.970 & 0.976 & \textbf{0.986}& \underline{0.985}& 0.978& 0.979& 0.981& 0.931& 0.952& 0.964& 0.943& 0.956& 0.962 \\
&ERGAS& 545.2 & 176.1& 142.5& 60.7& 52.4 & \textbf{43.7}& \underline{47.8}& 58.9& 57.5& 53.6& 121.6& 101.8& 80.1& 113.6& 111.8& 72.2 \\
& MSAM &0.836 & 0.536& 0.409& 0.115& 0.091& \underline{0.082}& \textbf{0.074}& 0.106& 0.108& 0.095& 0.332& 0.305& 0.251& 0.257& 0.281& 0.119\\
\Xhline{1\arrayrulewidth}
\multirow{4}{*}{\makecell[c]{Case\\(b)}}&MSPNR&19.87& 29.22& 31.16&  37.85& 40.01& \textbf{40.84}& \underline{40.50}& 40.07& 39.89& 40.45& 32.27& 35.08& 36.37& 33.74& 35.27& 37.17 \\
&MSSIM&0.412 & 0.849& 0.865& 0.962& 0.972& \textbf{0.982}& \underline{0.981}& 0.973& 0.972& 0.976& 0.928& 0.947& 0.958& 0.939& 0.953& 0.956 \\
&ERGAS&551.8& 176.6& 144.9& 176.0& 61.3&  \textbf{51.1}& \underline{55.9}& 61.8& 63.6& 57.0& 135.8& 109.3& 85.8& 114.6& 114.1& 76.0 \\
& MSAM &0.841 & 0.552& 0.412& 0.116& 0.093 & \underline{0.085}& \textbf{0.075}& 0.108& 0.110& 0.097& 0.332& 0.333& 0.269& 0.270& 0.302& 0.127\\
\Xhline{1\arrayrulewidth}
\multirow{4}{*}{\makecell[c]{Case\\(c)}}&MSPNR&11.83& 28.59& 29.03 & 22.23& 22.28 &22.57& 22.80& 34.64& 34.89& 36.58& 34.08& 36.21& \underline{37.43}& 34.19& 35.66& \textbf{37.46} \\
&MSSIM&0.147& 0.832& 0.841& 0.547& 0.548 & 0.552& 0.554& 0.915& 0.918& 0.940& 0.926& 0.948& \underline{0.956}& 0.928& 0.938& \textbf{0.957} \\
&ERGAS&1428&  220.9& 206.7& 458.6& 456.7& 449.2& 442.4& 187.4& 184.5& 100.2& 153.2& 95.4& \underline{81.3}& 108.5& 114.8& \textbf{80.6} \\
& MSAM &0.960& 0.316& 0.304& 0.368& 0.382&  0.348& 0.316& 0.308& 0.314& 0.232& 0.316& 0.262& 0.198& 0.184& \underline{0.162}& \textbf{0.131}\\
\Xhline{1\arrayrulewidth}
\multirow{4}{*}{\makecell[c]{Case\\(d)}}&MSPNR&11.85 & 28.34& 28.72& 21.88& 22.02& 22.19& 22.44& 34.46& 34.57& 36.34& 33.82& 35.97& \underline{37.06}& 33.87& 35.22& \textbf{37.10} \\
&MSSIM&0.143& 0.831& 0.839&  0.546& 0.546& 0.550& 0.552& 0.911& 0.914& 0.938& 0.921& 0.944& \underline{0.954}& 0.923& 0.931& \textbf{0.955} \\
&ERGAS&1420& 224.3& 210.2& 462.3& 461.2 & 454.6& 451.4& 193.8& 192.9& 104.5& 159.3& 99.3& \underline{85.8}& 112.6& 119.3& \textbf{84.6} \\
& MSAM &0.962 & 0.332& 0.326& 0.387& 0.395 &0.364& 0.342& 0.332& 0.346& 0.248& 0.343& 0.289& \textbf{0.216}& 0.212& 0.196& \textbf{0.158}\\
\Xhline{1\arrayrulewidth}
\multirow{4}{*}{\makecell[c]{Case\\(e)}}&MSPNR&13.33& 27.36& 28.23& 23.18&  22.92& 23.45& 23.84& 33.77& 33.96& 35.21& 33.23& 34.69& \textbf{35.81}& 33.48& 33.40& \underline{35.54} \\
&MSSIM&0.177 & 0.792& 0.798& 0.556& 0.547& 0.568& 0.572& 0.908& 0.911& 0.928& 0.900& 0.925& \textbf{0.941}& 0.914& 0.912& \underline{0.937} \\
&ERGAS&1208& 252.1& 240.2& 422.6& 437.2 & 412.6& 403.2& 132.2& 136.4& 103.4& 142.3& 112.3& \underline{96.5}& 116.8& 137.8& \textbf{94.8} \\
& MSAM &0.916 & 0.346& 0.338 & 0.373& 0.389& 0.372& 0.368& 0.316& 0.311& 0.231& 0.299& 0.276& \underline{0.278}& 0.257& 0.328& \textbf{0.164}\\
\Xhline{1\arrayrulewidth}
\multirow{4}{*}{\makecell[c]{Case\\(f)}}&MSPNR&13.21 &26.93& 28.11 & 23.06& 22.87 & 23.18& 23.45& 33.63& 33.82& 35.12& 33.12& 34.46& \textbf{35.53}& 33.20& 33.15& \underline{35.31} \\
&MSSIM&0.173& 0.784& 0.747& 0.594& 0.589& 0.605& 0.611& 0.907& 0.909& 0.927& 0.898& 0.923& \textbf{0.938}& 0.911& 0.907& \underline{0.934} \\
&ERGAS&1223& 263.2& 254.4& 425.6& 441.8 & 418.6& 422.1& 135.6& 138.1& 104.2& 145.6& 114.8& \underline{97.9}& 122.4& 145.1& \textbf{96.5} \\
& MSAM &0.926& 0.365& 0.352& 0.391& 0.401& 0.383& 0.382& 0.323& 0.321& 0.239& 0.306& 0.292& 0.281& \underline{0.266}& 0.347& \textbf{0.169}\\
\Xhline{1\arrayrulewidth}
\hline
\hline
\Xhline{1\arrayrulewidth}
 \multicolumn{2}{c|}{ Mean Times(s)}& & 13.9& 40.5& 46.2& 1586& 1111& 197.1& \underline{10.95}& 11.12& \textbf{10.72}& 201.8& 250.6& 44.9& 80.5& 149.3& 16.5\\
\Xhline{1\arrayrulewidth}
\Xhline{1pt}
\end{tabular}
\vspace{-0.2cm}
\end{table*}

\begin{figure*}[!]
		\centering
		\vspace{-5mm}
		\includegraphics[scale=0.43]{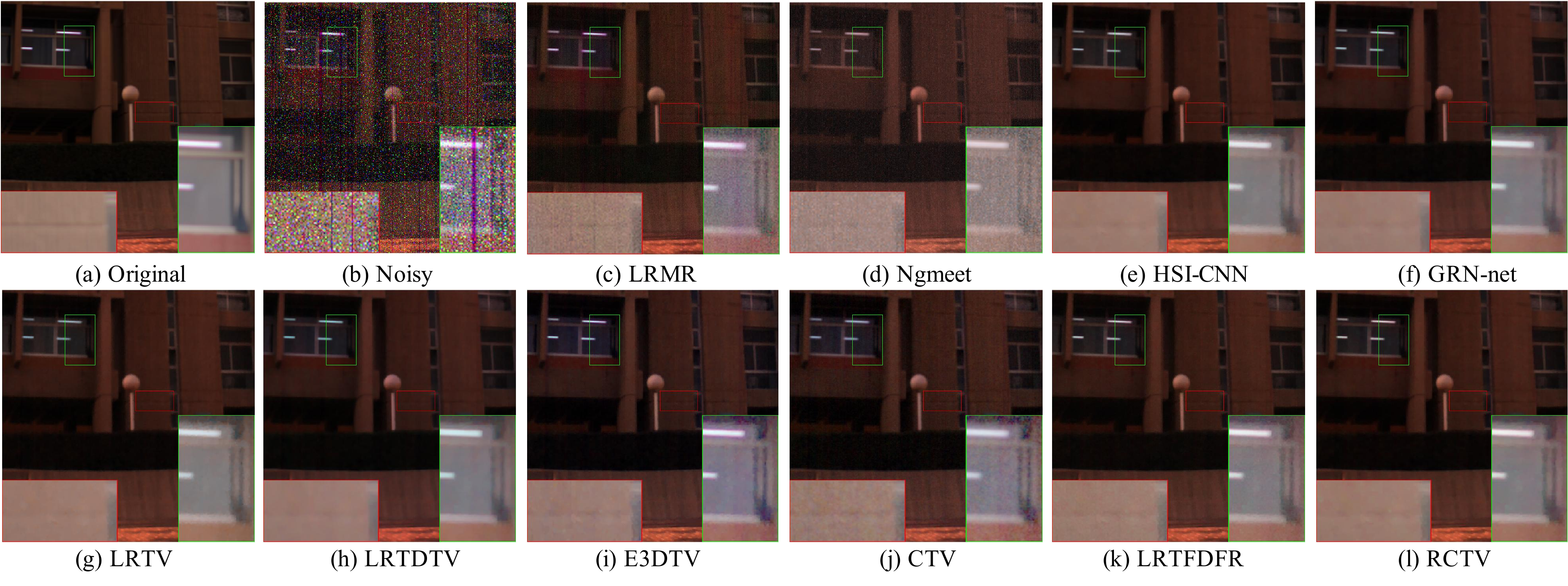}
		\vspace{-3mm}
		\caption{Recovered images of all competing methods with bands 23-13-4 as R-G-B.  (a) The simulated ICVL image. (b) The noisy images under case (f). (c-h) The recovered images obtained by all the competing methods, with a demarcated zoomed in 2.5 times for easy observation.}
		\label{icvl_3}
	\end{figure*}
	
	\begin{figure*}[!]
		\centering
		\vspace{-4mm}
		\includegraphics[scale=0.42]{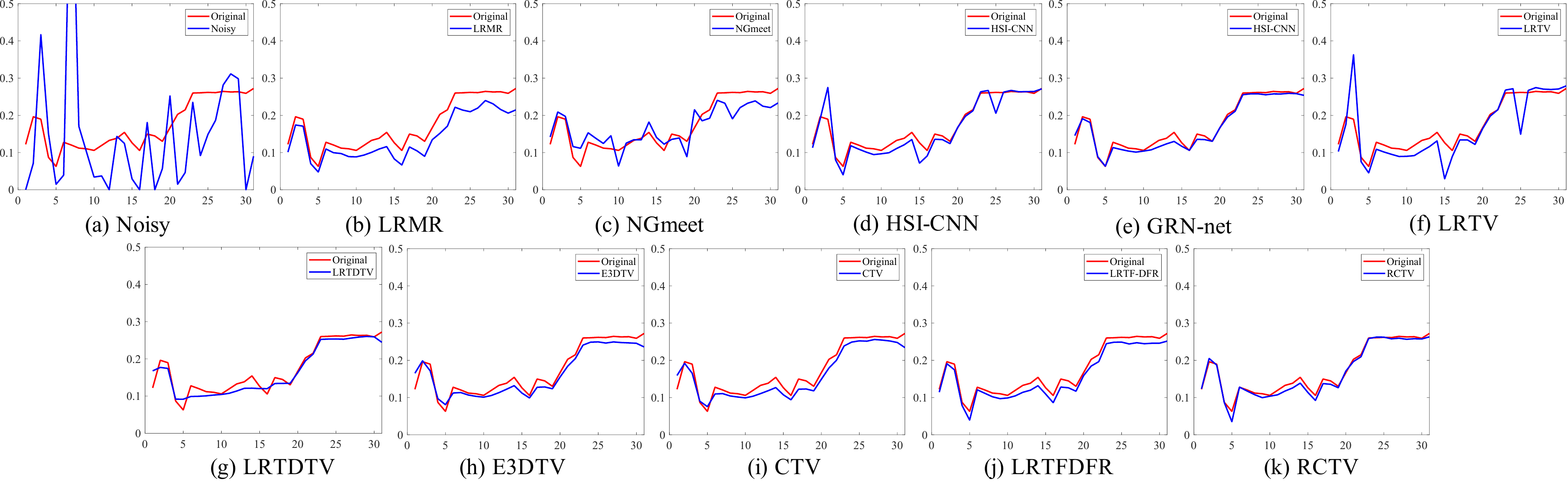}
		\vspace{-3mm}
		\caption{The spectral signatures of point (200, 300) in this ICVL dataset under case (f) before and after denoising by different methods from top to bottom, respectively.}
		\label{icvl_3_s}
	\end{figure*}

%\begin{figure*}[!]
%		\centering
%		\includegraphics[scale=0.48]{PIC/icvl_1.pdf}
%		\caption{Recovered images of all competing methods with bands 23-13-4 as R-G-B.  (a) The simulated ICVL image. (b) The noise images under case f). (c-h) The results were obtained by all comparison methods, with a demarcated zoomed in 2.5 times for easy observation.}
%		\label{icvl_1}
%	\end{figure*}
%
%	\begin{figure*}[!]
%		\centering
%		\includegraphics[scale=0.46]{PIC/icvl_1_s.pdf}
%		\caption{The spectral signatures of point (200, 300) in this ICVL dataset under case f) before and after denoising by different methods from top to bottle, respectively}
%		\label{icvl_1_s}
%	\end{figure*}

\begin{figure*}[!]
		\centering
		\vspace{-4mm}
		\includegraphics[scale=0.43]{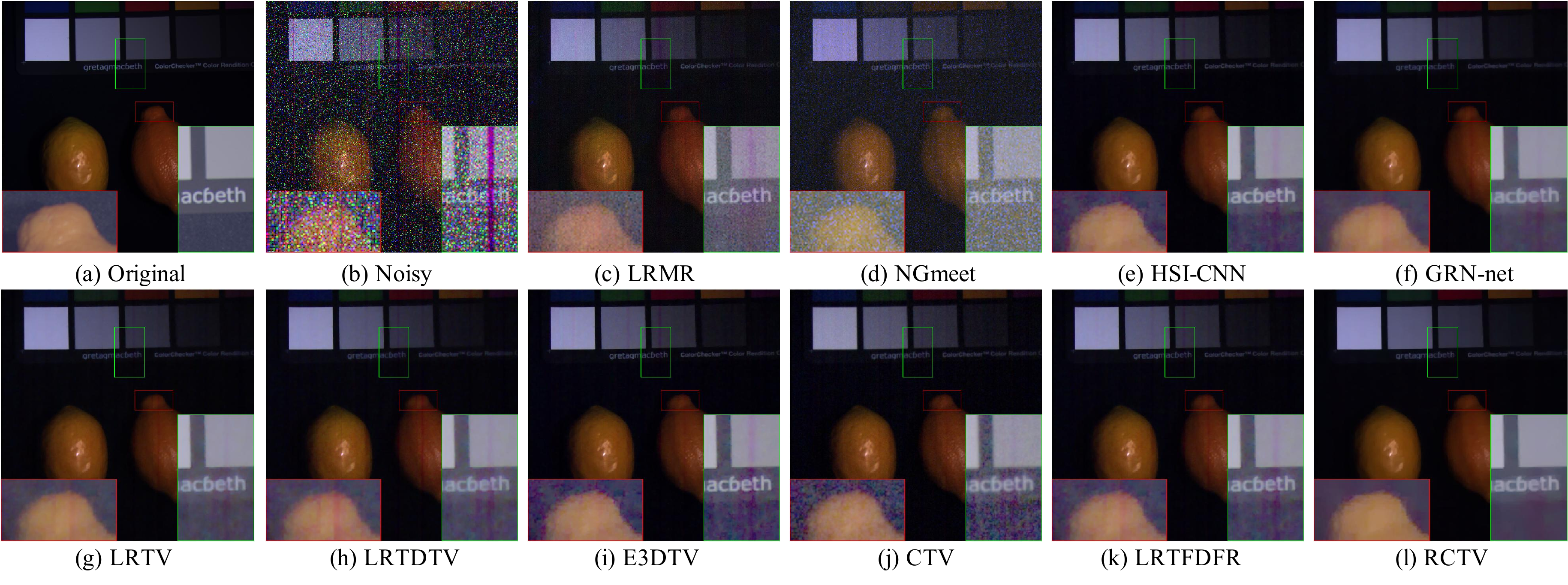}
		\vspace{-3mm}
		\caption{Recovered images of all competing methods with bands 23-13-4 as R-G-B.  (a) The simulated CAVE image. (b) The noisy images under case (f). (c-h) The recovered images obtained by all the competing methods, with a demarcated zoomed in 2.5 times for easy observation.}
		\label{cave_4}
	\end{figure*}
	
	\begin{figure*}[!]
		\centering
		\vspace{-4mm}
		\includegraphics[scale=0.42]{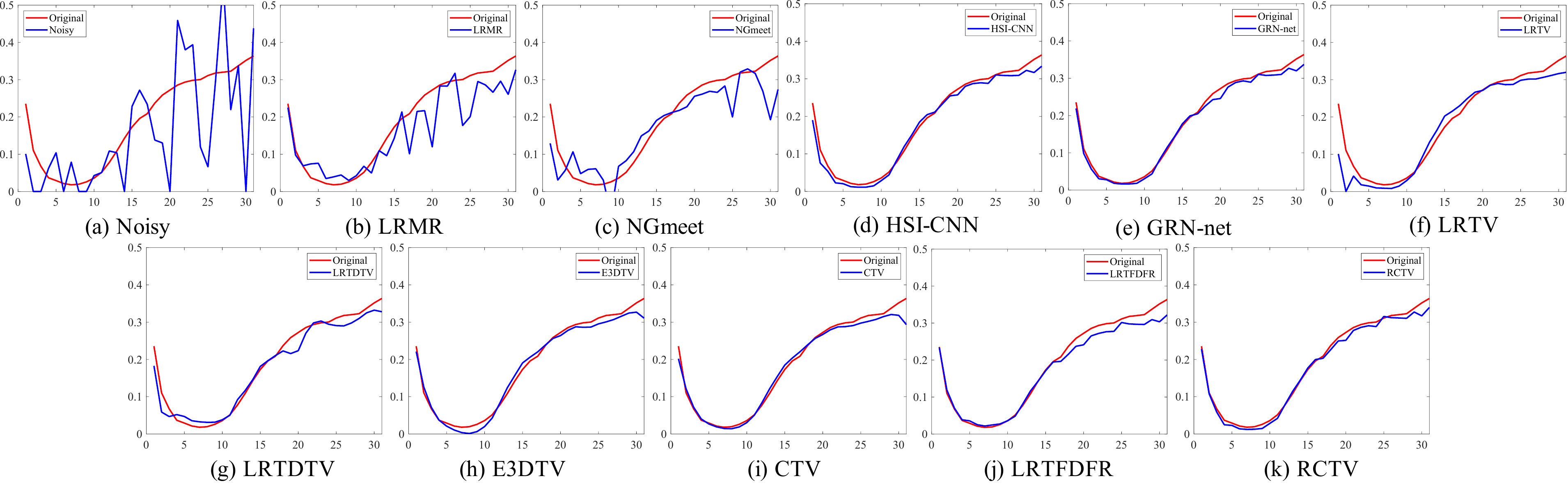}
		\vspace{-4mm}
		\caption{The spectral signatures of point (200, 200) in this CAVE dataset under case (f) before and after denoising by different methods from top to bottom, respectively. }
		\label{cave_4_s}
	\end{figure*}

\subsubsection{Denoising Results for MSI data}
We show the results of all competing methods on two MSI datasets.

The quantitative results are given in Tables \ref{icvl_table}  and \ref{cave_table}, where each value is the average of ten test images. First of all, we find that the performances of RCTV are relatively worse than that of LLRT and NGmeet in case (a) and case (b). The reason behind this is that NLS prior is very effective to remove Gaussian noise, and we have known that NGmeet and LLRT are two leading models for Gaussian noise removal. Although our method is inferior to NGmeet in removing Gaussian noise, it still achieves the best results among all the state-of-the-art methods based on LS prior. However, stripes, deadlines, and impulse noises are often encountered in practical situations, so we further conduct denoising experiments for complex noise from case (c) to case (f). We can easily find that when the noise becomes complex, the performance of these models based on NLS prior begin to drop significantly due to the inaccuracy of finding similar blocks, while our RCTV model can still maintain stable and great denoising performance. Specifically, the evaluation metrics of RCTV are much higher than other state-of-the-art methods, such as LRTDTV, LRTFDFR, E3DTV. These methods have been verified to be very effective in dealing with complex noise. In addition, we also noticed that although the deep learning-based methods did not achieve the best performance, considering its testing time and denoising performance, deep learning methods still release great potential in the HSI denoising task.

From the above analysis, we have already known that our RCTV model can get the best denoising performance in most noise situations. As we explained earlier, mining the priors of the representative coefficient instead of the original data can also significantly reduce the runtime. From the last row of Table \ref{icvl_table} and Table \ref{cave_table}, we can observe that the runtime of our RCTV model is the lowest among all models based on L and LS priors, which means that the $ \U $-based model can indeed greatly reduce the running time. Further, we compare the denoising time of our RCTV with that of the models based on pure L prior. Generally, models based on pure L prior have a short denoising time, but the denoising performance is worse than models based on multiple priors. However, we can find that the time of our proposed RCTV is lower than that of the models based on L prior, such as TDL \cite{peng2014decomposable} and LRMR \cite{zhang2013hyperspectral}. Therefore, our RCTV can greatly reduce denoising time while obtaining better performance. Surprisingly, the runtime of our RCTV model is even comparable with the test time of the deep learning methods.

Furthermore, we select two images to show some false color images of all competing methods. Specifically, we not only show the restored images of different methods, but also provide spectral signature curves obtained by different denoising methods in Fig. \ref{icvl_3}-Fig. \ref{cave_4_s}. The reason behind this is that the spectral signature curves are very important for the HSI denoising task. Each point in the spatial space represents a type of substance, and each type of substance has a specific spectral signature, which reflects the physical properties of this type of material. Therefore, HSI denoising is not only to remove the noise more fully on a single image, but also to fully maintain the spectral curves of the substances. Aiming at a better visual comparison, we mark the same subregion of each subfigure in Fig. \ref{icvl_3}, and Fig. \ref{cave_4} and then enlarge it 2.5 times. Several observations can be easily made from these figures. Firstly, all the compared methods can somewhat remove such mixed noises. Secondly, the proposed RCTV method outperforms all the compared methods. Precisely, RCTV can effectively remove the impulse and stripe noises while retaining the spatial texture of the original HSI, and obtaining better image color fidelity. Additionally, RCTV can better maintain the spectral signature curve of the substance. In Fig. \ref{icvl_3_s} and Fig. \ref{cave_4_s}, we can clearly see that the spectral signature curve of RCTV is the closest to the groundtruth.

Therefore, based on the above analysis, it is not difficult to conclude that RCTV can not only greatly reduce the running time of denoising, but also achieve the best performance in the complex noise case.

\begin{figure*}[!]
		\centering
		\includegraphics[scale=0.52]{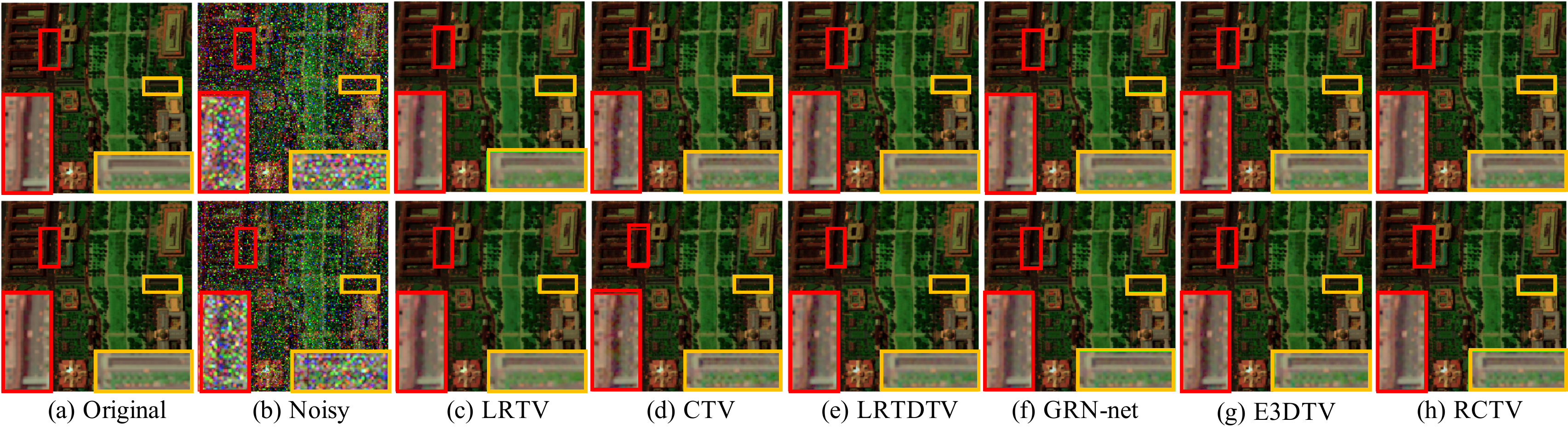}
		\vspace{-3mm}
		\caption{Recovered images of all competing methods with bands 6-105-155 as R-G-B.  (a) The simulated DC mall image. (b) The noisy images under case (c) and (e) from top to bottom. (c-h) The recovered images by all the competing methods, with a demarcated zoomed in 2.5 times for easy observation.}
		\label{test_dcmall}
	\end{figure*}
	
		\begin{figure*}[!]
		\centering
		\includegraphics[scale=0.45]{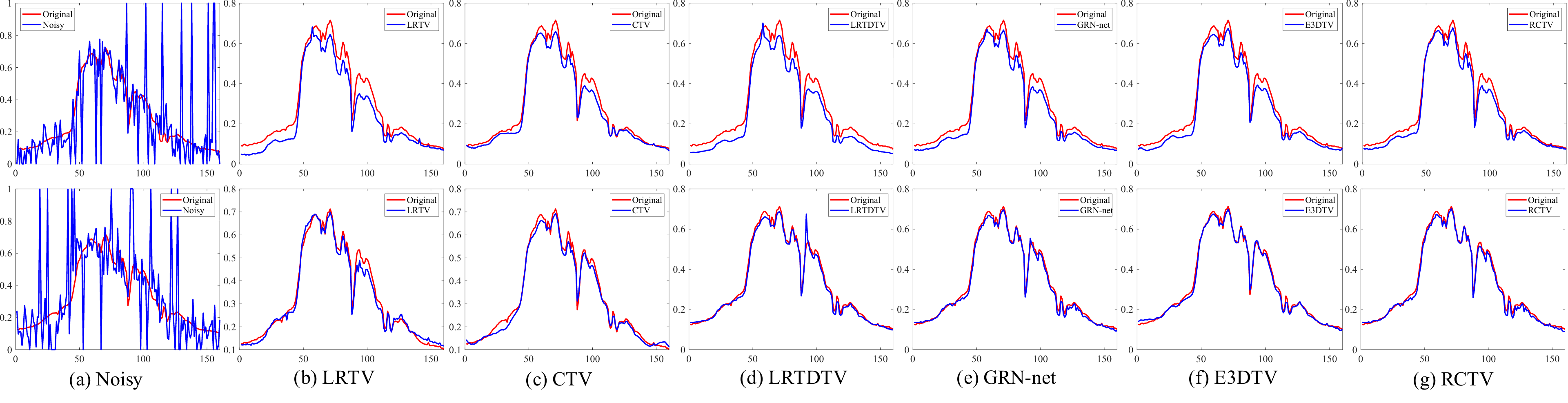}
		\vspace{-3mm}
		\caption{The spectral signatures of point (100, 100) in DC mall dataset under case (c) and point (120, 120) in DC mall dataset under case (e) before and after denoising by different methods from top to bottom, respectively.}
		\label{test_dcmall_spectral}
	\end{figure*}

\subsubsection{Denoising Results for HSI Data}
The previous subsection shows the denoising results of our method on MSI data. In this chapter, we directly conduct simulation experiments on HSI data. Compared with MSI, the band number of HSI is larger and thus the low-rankness of HSI is stronger. All model-based methods explicitly exploit L prior, so model-based methods perform relatively better than data-based methods. Since our method is based on mining the L and LS prior, the restoration performance of our RCTV will be more prominent on this HSI data.

The quantitative comparison, the visual restored image of all competing methods, and the restored signature curves are provided in Table \ref{dcmall_table}, Fig. \ref{test_dcmall} and Fig. \ref{test_dcmall_spectral}, respectively. From these tables and figures, we can easily observe similar results to MSI experiments. In addition, with the enhancement of the low-rankness of HSI data, the evaluation metrics obtained by some traditional methods such as LRTDTV and E3DTV on complex noise will be better than the denoising performance of deep learning-based approaches. Indeed, the performance of our method can also be further enhanced.

\begin{table*}[htbp]
\renewcommand{\arraystretch}{1.15}
\setlength\tabcolsep{3.0pt}
\footnotesize
  \caption{Quantitative comparison of all competing methods under different levels of noises in the \textbf{DC mall} dataset. The best and second results are highlighted in bold and \underline{underline}, respectively.}
  \label{dcmall_table}
  \setlength{\abovecaptionskip}{5pt}
  \setlength{\belowcaptionskip}{5pt}
  \centering
  \vspace{-0.2cm}
\begin{tabular}{l|c|c||c|c|c||c|c|c||c|c|c||c|c|c|c|c|c}
     \Xhline{1pt}
     	\multirow{3}{*}{\makecell[c]{Noise\\Types}} &\multirow{3}{*}{Metric}& \multirow{3}{*}{Noisy}& \multicolumn{3}{c||}{ \textbf{L}} & \multicolumn{3}{c||}{ \textbf{L \& NLS}} & \multicolumn{3}{c||}{ \textbf{DL}}  & \multicolumn{6}{c}{ \textbf{L\& LS}}  \\
     	\cline{4-18}
     	& & &WNNM  &\multirow{2}{*}{LRMR} & \multirow{2}{*}{TDL} &\multirow{2}{*}{KBR}&\multirow{2}{*}{LLRT}& NG-  &HSI- &HSI- &GRN & \multirow{2}{*}{LRTV}  &LRTD &\multirow{2}{*}{E3DTV}  & CTV- & LRTF-& \multirow{2}{*}{\textbf{RCTV}}\\
		& & &-RPCA & & & & & Meet & CNN & DeNet & -Net & & -TV & & RPCA &DFR \\
		\Xhline{1pt}
\multirow{4}{*}{\makecell[c]{Case\\(a)}} & MPSNR &20.00 & 33.00& 33.77& 35.53& 35.61& 36.42& \textbf{37.98}& 36.28& 36.15& 36.96 & 35.16& 35.23& 36.56& 35.31 &36.88& \underline{37.21}\\
& MSSIM&0.515& 0.957& 0.958& 0.971& 0.972& 0.978& \textbf{0.983}& 0.967& 0.965& 0.978& 0.964& 0.964& 0.976&0.971 & 0.978&\underline{0.979} \\
& ERGAS& 375.9 & 83.4& 74.33& 60.37& 62.88& 59.94& \textbf{45.24}& 59.43& 58.37& 51.46& 62.98& 61.83& 55.74& 63.75&52.86& \underline{49.90} \\
& MSAM&0.475& 0.110& 0.102& 0.079& 0.066& 0.070& \textbf{0.059}& 0.071& 0.069& 0.066& 0.093& 0.081& 0.075&0.077&\underline{0.065} & 0.069 \\
\Xhline{1\arrayrulewidth}
\multirow{4}{*}{\makecell[c]{Case\\(b)}}& MPSNR&19.87 & 32.93& 33.55 & 34.51& 34.82 & 35.47& \textbf{37.06}& 35.58& 35.17& 36.58& 34.18& 34.33& 35.57& 35.15& 36.46 & \underline{36.65} \\
& MSSIM&0.509 &  0.957& 0.957& 0.965& 0.966& 0.972& \textbf{0.980}& 0.958& 0.945& 0.976& 0.953& 0.958& 0.970&0.971 & 0.976 & \underline{0.977}\\
& ERGAS& 381.1 & 83.92& 76.63& 75.52& 71.38 & 67.21& \textbf{52.93}& 62.24& 64.28& 54.68& 76.38& 68.90& 61.19&64.76 & 55.19 & \underline{53.58}\\
& MSAM&0.481 & 0.110& 0.101& 0.085& 0.077& 0.076& \textbf{0.066}& 0.075& 0.073& \underline{0.068}& 0.092& 0.083& 0.080& 0.078& 0.066 & 0.076 \\
\Xhline{1\arrayrulewidth}
\multirow{4}{*}{\makecell[c]{Case\\(c)}} & MPSNR&12.38 & 33.49& 32.77&  23.73& 25.34 & 25.48& 26.46& 32.62& 32.87& 34.89 & 34.46& 35.41& \underline{36.35}& 35.49& 35.62& \textbf{37.02} \\
& MSSIM& 0.221 & 0.963& 0.960& 0.801& 0.830 & 0.832& 0.846& 0.956& 0.958& 0.962& 0.956& 0.967& \underline{0.975}& 0.970&0.969 &\textbf{0.978} \\
& ERGAS& 925.1 & 80.39& 86.41& 246.7& 168.3 & 174.8& 161.6& 77.3& 74.5& 65.8& 97.59& 63.98& \underline{58.49}& 61.86& 61.14& \textbf{51.01} \\
& MSAM& 0.733 & 0.131& 0.092&  0.189& 0.183 & 0.181& 0.174& 0.092& 0.095& 0.075& 0.131& 0.082& \underline{0.075}& 0.081& 0.080 &\textbf{0.069}\\
\Xhline{1\arrayrulewidth}
\multirow{4}{*}{\makecell[c]{Case\\(d)}} & MPSNR& 12.37 & 33.4& 32.61 & 23.8& 25.12& 25.36& 26.02& 32.18& 32.53& 34.68& 34.33& 35.23& \underline{36.08}&35.62 & 35.65&\textbf{36.85} \\
& MSSIM& 0.219 &  0.963& 0.957 & 0.800& 0.822 &0.825& 0.843& 0.953& 0.954& 0.962& 0.955& 0.966& \underline{0.973}& 0.971&0.969& \textbf{0.978} \\
& ERGAS&926.3 & 80.39& 96.41& 246.7& 172.1 & 176.2& 162.4& 80.4& 78.6& 63.2& 97.59& 63.98& \underline{58.49}& 61.36&62.97& \textbf{52.84} \\
& MSAM&0.736 & 0.132& 0.094& 0.190& 0.188 & 0.185& 0.174& 0.094& 0.097& 0.077& 0.134& 0.084& \underline{0.080}& 0.082& 0.083&\textbf{0.072} \\
\Xhline{1\arrayrulewidth}
\multirow{4}{*}{\makecell[c]{Case\\(e)}}& MPSNR&13.73 & 32.72& 32.83 & 25.64& 26.84 & 26.42& 27.34& 33.52& 33.36& 34.54& 33.37& 34.25& \underline{35.42}& 34.47&34.49& \textbf{35.97} \\
& MSSIM&0.277& 0.955& 0.956& 0.842& 0.857 &0.852& 0.868& 0.956& 0.958& 0.962& 0.946& 0.958& \underline{0.969}&0.962 &0.964& \textbf{0.972} \\
& ERGAS&805.9 & 85.48& 83.4& 203.7& 184.3 & 189.4& 172.6& 75.6& 74.8& 63.82& 99.25& 70.4& \underline{62.27}&69.46 & 68.43& \textbf{59.15} \\
& MSAM&0.688 & 0.137& 0.103& 0.178& 0.163 & 0.166&0.161& 0.097& 0.096& 0.082& 0.129&0.089& \underline{0.084}&0.091&0.100& \textbf{0.081} \\
\Xhline{1\arrayrulewidth}
\multirow{4}{*}{\makecell[c]{Case\\(f)}}& MPSNR&13.53 & 32.76& 33.82 & 25.56& 26.66 & 26.26& 27.18& 33.48& 33.25& 34.56& 33.43& 34.34& \underline{35.23}& 34.50& 34.57& \textbf{35.64} \\
& MSSIM& 0.266& 0.956& 0.965& 0.838& 0.853 & 0.848& 0.864& 0.964& 0.964& \underline{0.967}& 0.944& 0.957& \underline{0.967}&0.963 & 0.963 &\textbf{0.971} \\
& ERGAS& 817.9 & 85.78& 108.2& 208.5& 186.5 & 191.4& 174.8& 115.6& 114.2& 68.4& 105.8& 71.16& \underline{63.58}&68.42 &68.61& \textbf{59.65} \\
& MSAM&0.698 & 0.148& 0.121& 0.181& 0.166 & 0.171& 0.163& 0.137& 0.132& 0.124& 0.136& 0.112& 0.109& \underline{0.104}& 0.116& \textbf{0.095}\\
\Xhline{1\arrayrulewidth}
\hline
\hline
\Xhline{1\arrayrulewidth}
\multicolumn{2}{c|}{ Mean Times(s)}& & 9.10& 22.42& 12.83& 1721 & 1546& 34.23& 2.75& \underline{2.74}& \textbf{2.61}& 76.25& 114.2& 38.56& 58.81& 38.92& 5.82\\
\Xhline{1pt}
\end{tabular}
\vspace{-0.2cm}
\end{table*}

\subsection{Real HSI Denoising}
Two real-world HSI data sets used in \cite{wang2017hyperspectral, peng2020enhanced} are used in our experiments, i.e., the Hyperspectral Digital Imagery Collection Experiment (HYDICE) Urban data set with size $ 307\times 307 \times 210 $, and the Hyperspectral Digital Imagery Collection Experiment (HYDICE) Terrian data set with size $ 300 \times 300 \times 210 $. For these two data sets, some bands are seriously polluted by the atmosphere and water and corrupted by complex noises (e.g., deadline, stripe, sparse and Gaussian noise), thus it is a big challenge to remove the noise.

In this experiment, the competing methods include WNNM \cite{gu2014weighted, gu2017weighted}, LRMR \cite{zhang2013hyperspectral}, NGmeet \cite{he2020non}, LRTV \cite {he2015total}, LRTDTV \cite{wang2017hyperspectral}, E3DTV \cite{peng2020enhanced}, CTV-RPCA \cite{peng2022exact}, LRTFDFR \cite{zheng2020double}, and two deep learning methods (i.e., HSICNN \cite{yuan2018hyperspectral}, GRN-net \cite{cao2021deep}).

\begin{figure*}[!]
		\centering
		\includegraphics[scale=0.57]{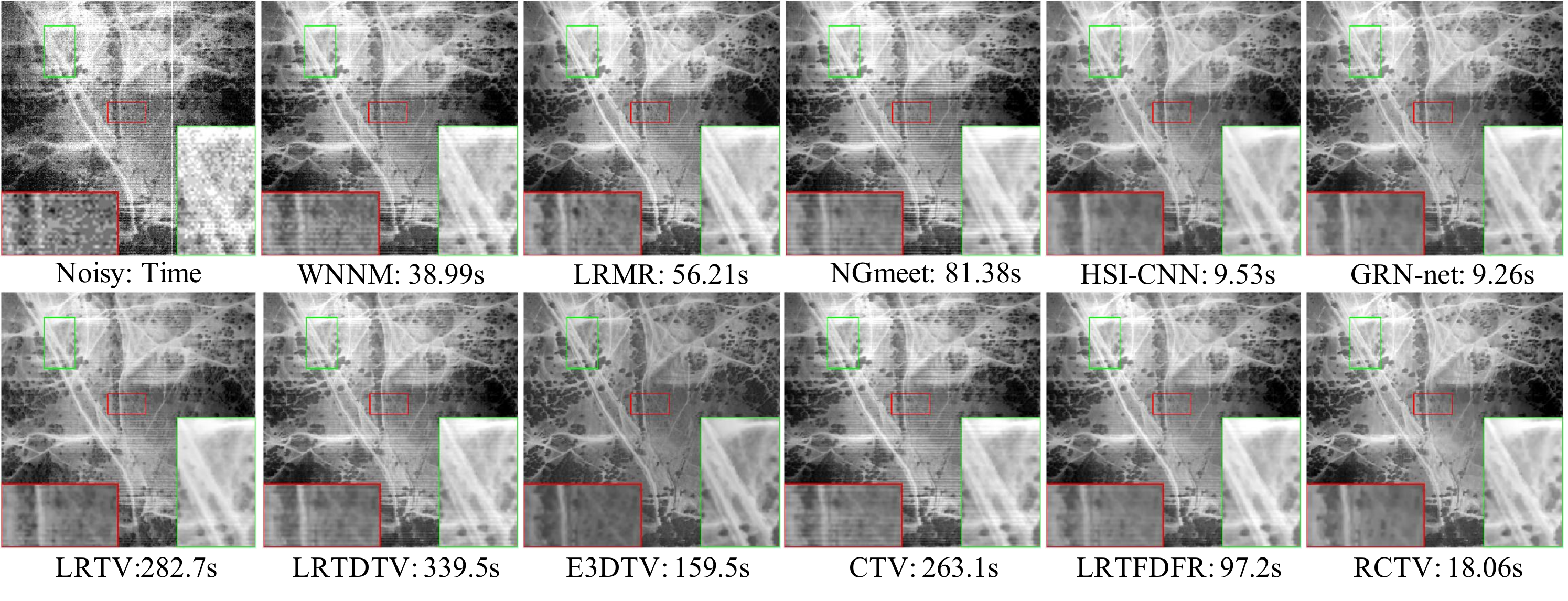}
		\vspace{-3mm}
		\caption{Recovered images of all competing methods at band 139 of Terrian data.}
		\label{terrian_result}
	\end{figure*}
	
	\begin{figure*}[!]
		\centering
		\vspace{-4mm}
		\includegraphics[scale=0.57]{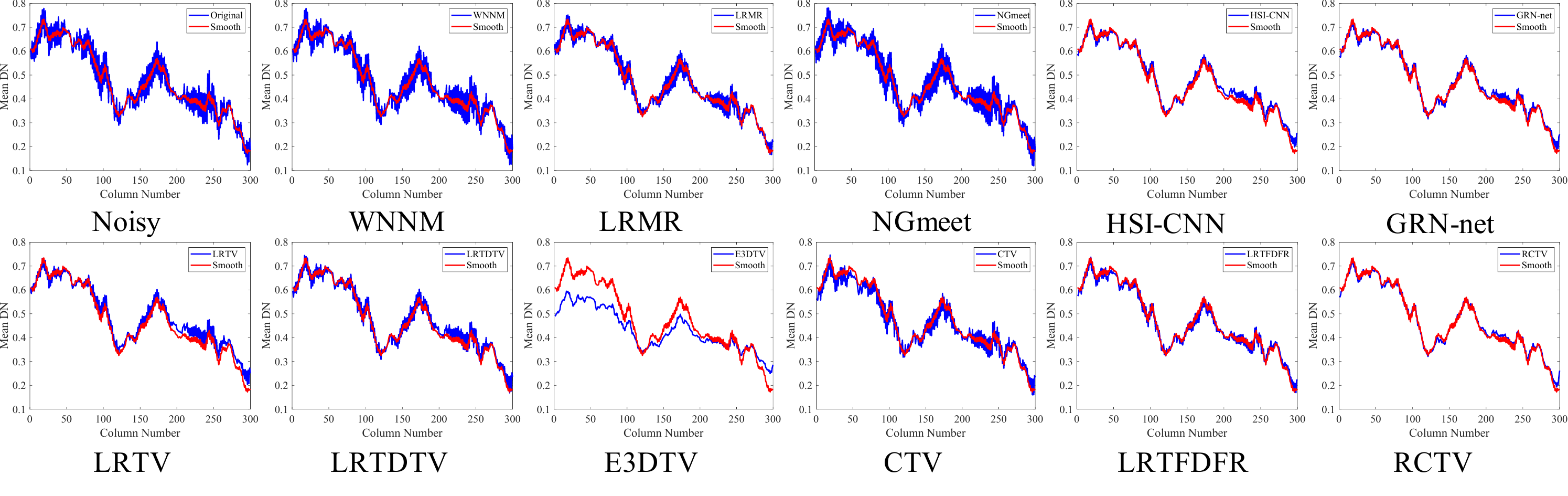}
		\vspace{-3mm}
		\caption{The vertical mean profiles of all competing methods at band 104 of Terrian data.}
		\label{terrian_result_spectral}
	\end{figure*}
	
	\begin{figure*}[!]
		\centering
		\vspace{-4mm}
		\includegraphics[scale=0.57]{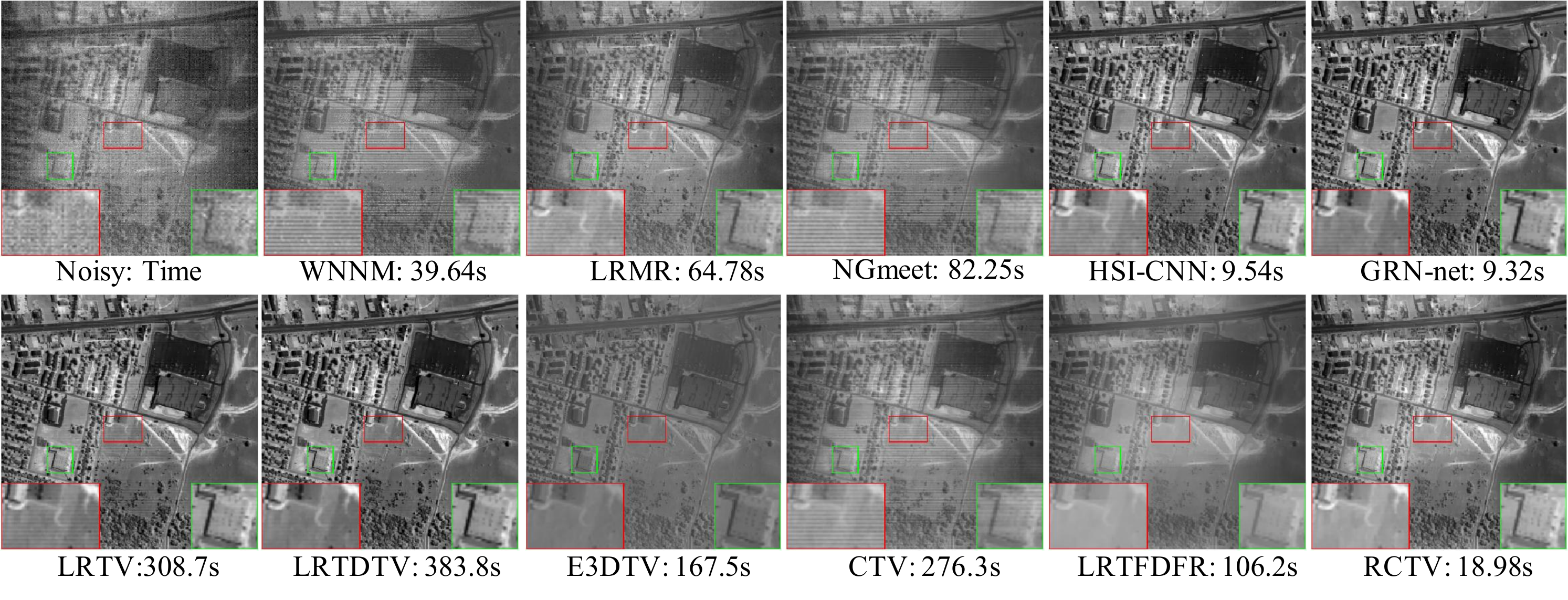}
		\vspace{-3mm}
		\caption{Recovered images of all the competing methods at band 104 of Urban data.}
		\label{urban_result}
	\end{figure*}
	
		\begin{figure*}[!]
		\centering
		\vspace{-3mm}
		\includegraphics[scale=0.57]{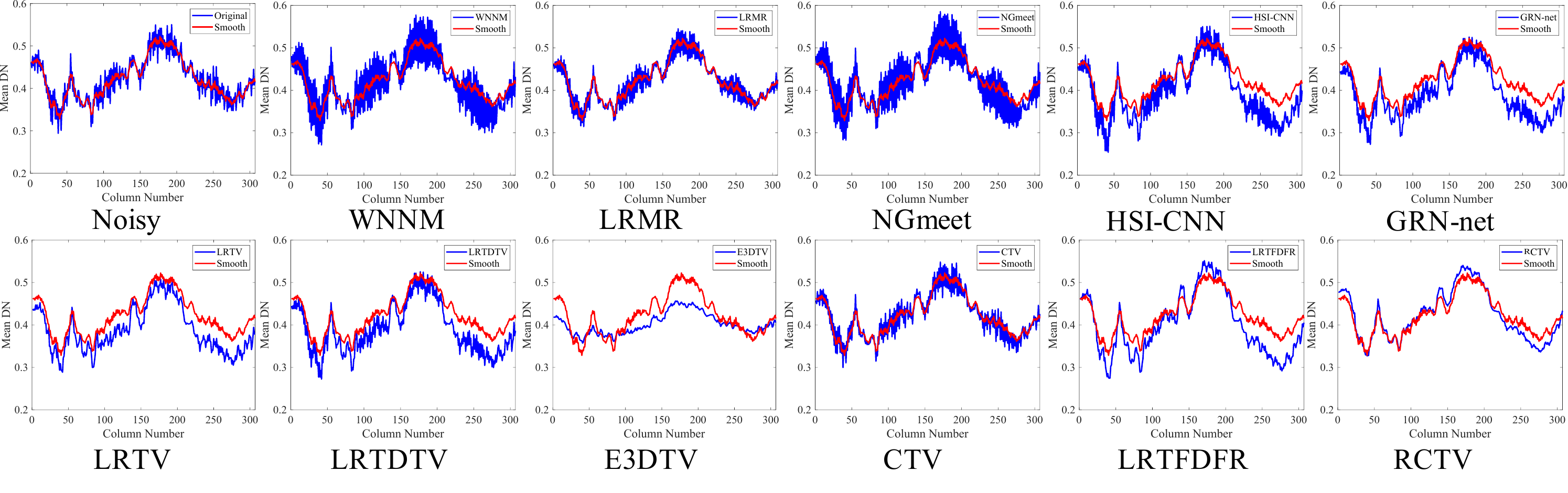}
		\vspace{-3mm}
		\caption{ The vertical mean profiles of the all competing methods at band 104 of  Urban data.}
		\label{urban_result_spectral}
	\end{figure*}

Since these are real HSI data, we cannot get the noiseless image and get evaluation metrics to judge the effect of competing methods. To make the comparison more comprehensive and reliable, we select images with a moderate degree of pollution, and show the denoised image and the column mean curve of the denoised image to assist in judging the denoising effect of the competing methods. We choose moderately polluted images for presentation because results based on heavily polluted images are often unconvincing. The reason for choosing to display the column mean curve of the repaired image is that the visual observation is often biased, and we can better measure the quality of the repair result based on the intrinsic properties of the image, such as local smoothness. Since the image has local smoothness, we can calculate the mean value of each column of the image and get the column mean curve, then the  column mean curve should also have a certain degree of smoothness. In addition, for the case where the noise level is not extremely serious, the column mean curves of noisy image and denoised image should be relatively close.

For a fair comparison, we provide the denoised images and column mean curves of all competing methods for Terrian data at band 139 in Fig. \ref{terrian_result}-\ref{terrian_result_spectral}, and Urban data at band 104 in Fig. \ref{urban_result}-\ref{urban_result_spectral}. From these figures, we can observe that all competing methods can remove the mixed noise to some extent. But for stripe noise, only our method and two deep learning methods can remove the noise better. Compared with the column mean curve of all methods, only our method can better maintain the trend of the column mean curve of the original image. Therefore our method achieves the best denoising performance. In addition, we also list the runtime of all competing methods. As can be seen from Fig. \ref{terrian_result} and Fig. \ref{urban_result}, the runtime of our proposed RCTV model is second fastest compared with all the competing methods, and almost comparable with the test time of the deep learning-based methods.
	
\subsection{Discussion}
\subsubsection{Parameter Analysis}
There are four parameters in RCTV model that need to be given in advance, i.e., the trade-off parameters $ R $, $ \tau $, $ \beta $ and $ \lambda $.

\begin{figure}[!]
		\centering
		\includegraphics[scale=0.6]{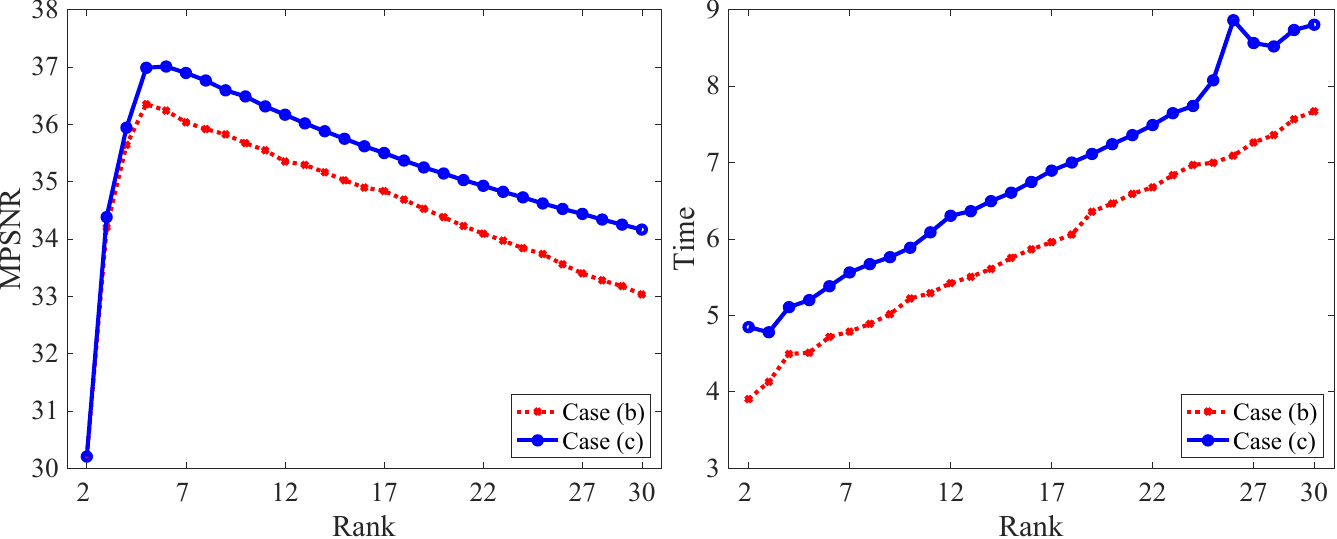}
		\vspace{-4mm}
		\caption{ Sensitivity analysis of the rank $ R $ under case (b) and case (c) on DC mall data.}
		\label{Rank_parameter}
	\end{figure}

The rank $ R $ mainly characterizes the spectral global correlation of HSIs.
According to the analysis in Table \ref{model_complexity}, we have known that the time complexity of RCTV is related to $ R $. Therefore, we provide the sensitivity analysis of the rank  $R $ and runtime of RCTV under different $ R $ of Fig. \ref{Rank_parameter}. From the left sub-figure in Fig. \ref{Rank_parameter}, it can be seen that the proposed RCTV exhibits stable and superior performance within a certain range of rank under case (b) and case (c) on DC mall data. From the right sub-figure of  Fig. \ref{Rank_parameter}, it is easily seen that the runtime is nearly positively correlated with $ R $, which is consistent with the time comparison reported in Table \ref{model_complexity}. To balance the good denoising result and denoising time, the rank of all the simulation experiments in this paper is selected from 6 to 10. As for real HSI data, the rank is estimated by the well-known HySime algorithm \cite{bioucas2008hyperspectral}.

\begin{figure}[!]
		\centering
		\includegraphics[scale=0.4]{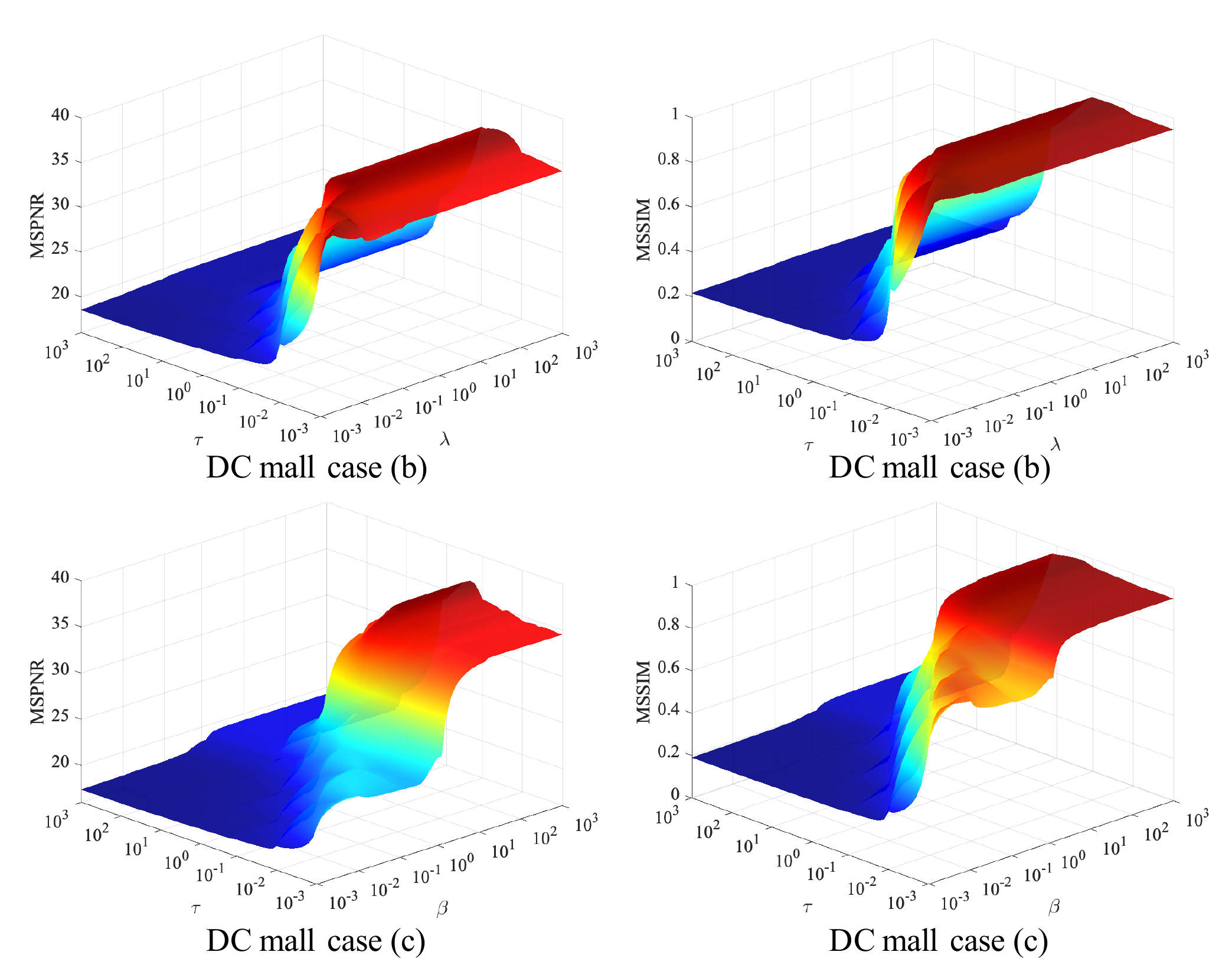}
		\vspace{-4mm}
		\caption{ Sensitivity analysis of the parameters $ \tau $ and $ \lambda $ under Case (b), and sensitivity analysis of the parameters $ \tau $ and $ \beta $ under Case (c).}
		\label{lambda_beta}
	\end{figure}

At first glance, the objective function in model (\ref{alm_model}) contains three parameters, i.e., $\tau$, $\beta$ and $\lambda$. In fact, there is a certain proportional relationship between the three parameters. Using this relationship, we only need to fine-tune two parameters, or even one parameter to achieve a good denoising performance. Specifically, for the case where most noises obey Gaussian distribution, we can ignore the $ \ell_1 $ norm designed for the sparse noise. Then we can simply set $ \beta =1 $ and fine-tune $ \tau $. For the case where most noises obey laplacian distribution or mixed distribution, we can simply set $ \lambda =1 $, set $ \beta $ as a big value, and then fine-tune $ \tau $. Therefore, the performance of the RCTV model (\ref{alm_model}) is sensitive to $ \tau $, and is robust to $ \beta $ and $ \lambda $. To be more intuitive, we provide sensitivity analysis on the simulated DC mall data under Case (b) and (c) in Fig. \ref{lambda_beta} since they represent the Gaussian noise case and the mixed noise case, respectively. As observed, the performance of the proposed RCTV is robust to the parameters $ \lambda $ and $\beta$, and we can set $ \lambda$ or $ \beta $ with a large value. While it is sensitive to the parameter $ \tau $, we need to fine-tune $ \tau $ to get a good performance. Specifically, for Gaussian noise removal, i.e. cases (a) and (b), we fix $\beta = 1, \lambda=100$, and then finely choose $\tau$ from the set $ [0,2] $. For mixed noise removal, i.e, cases (c)-(f), we fix $ \lambda = 1, \beta = 50 $, and then finely choose $\tau$ from the set $ [0,2] $.

	\begin{figure}[!]
		\centering
		\includegraphics[scale=0.6]{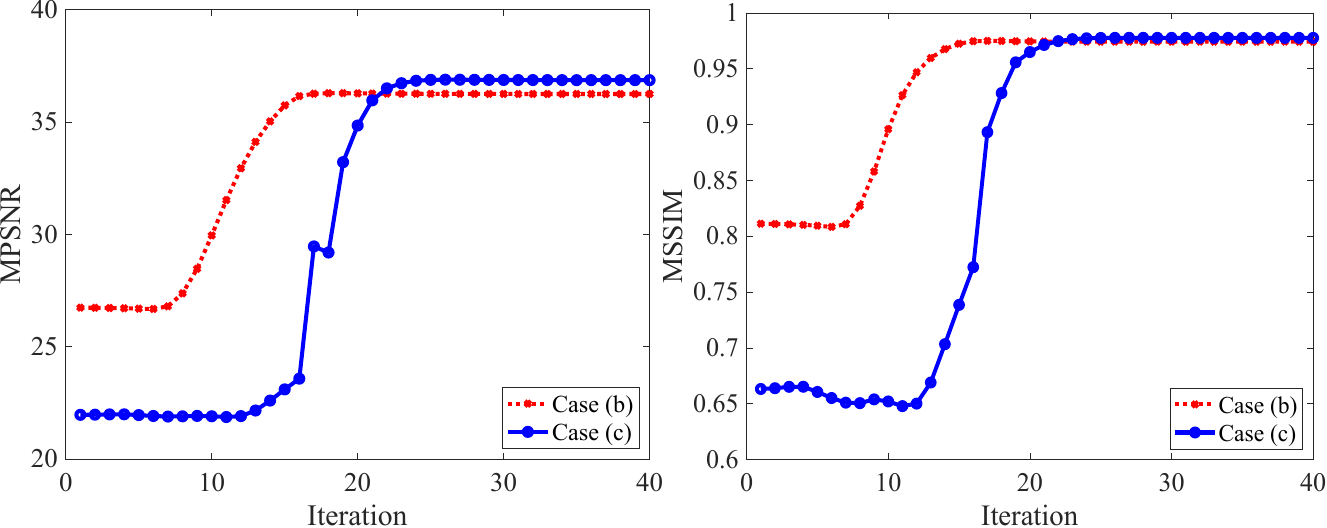}
		\vspace{-3mm}
		\caption{MPSNR and MSSIM value versus the iteration number of RCTV slover.}
		\label{iteration_rbtv}
	\end{figure}

\subsubsection{Convergence Analysis}  Since the RCTV model (\ref{alm_model}) is proposed based on LRMF, and LRMF is a non-convex framework, it is difficult to give a convergence theorem guarantee via the ADMM framework.  Instead, we numerically demonstrate the convergence. Fig. \ref{iteration_rbtv} presents the MPSNR and MSSIM values versus the iteration number of the RCTV solver. It can be observed that, as the number of iterations increases to a relatively large value, the relative changes of MPSNR and MSSIM converge to zero. This clearly illustrates the strong convergence of the proposed ADMM algorithm to solve the RCTV model, which further encourages us to utilize it for more practical applications.

\section{Conclusion}
\label{Conclusion_Part}
In this paper, we propose a simple, efficient and fast method for HSI mixed noise removal. Through our analysis, it is pointed out that the spatial information of the HSI can be transferred to its representative coefficients. By encoding the local smoothness of representative coefficients, we provide a new regularizer named RCTV that can simultaneously encode the global correlation and local smoothness of HSI. Since the band number of the representative coefficients matrix is far less than the band number of HSI data and the representative coefficients matrix is somewhat robust to noise, the model based on RCTV can not only greatly reduce the runtime of HSI denoising, but also further improve the performance. A series of simulated and real data experiments have been conducted to demonstrate the superior performance of the proposed method over some popular methods in terms of both the evaluation metrics and denoising runtime. 

In the future, we will focus more on mining the prior of representative coefficient matrices, solving the problem of parameter selection and recoverable theory. For mining the prior of representative coefficient matrices, we can utilize the deep neural network (DNN) to learn the prior of representative coefficient matrices from big data or propose other regularization terms to further improve performance and reduce runtime. Additionally, we will make more attempts to build recoverable theory and solve the parameter selection problem.
			
\bibliographystyle{IEEEtran}
\bibliography{mybibfile}	

\appendix
\label{appendix_theorem}
In this section, we give the proof of Theorem \ref{theorem_e}.
\begin{proof}
Write the SVD of $ \X \in \mathbb{R}^{MN\times B} $ of rank $ R $ as
\begin{equation}
\X = \hat{\U}\mathbf{\Sigma}\hat{\V}^{\textrm{T}} = \sum_{k=1}^B  \sigma_k \mathbf{u_k} \mathbf{v_k}^{\textrm{T}},
\end{equation}
where $ \sigma_k, (k=1,\cdots,R) $ are the positive singular values and $ \sigma_{R+1},\cdots, \sigma_{B}$ are zero since the rank of $ \X $ is $ R $, and $\hat{\U} =[\mathbf{u_1},\cdots,\mathbf{u_B} ]$, $\hat{\V} =[\mathbf{v_1},\cdots,\mathbf{v_B} ]$ are the matrices of left- and right-singular vectors. Combine $ \hat{\U}$ and $ \mathbf{\Sigma}  $ into one, still record as $ \hat{\U}$.  Since only the first $ R $ singular values are non-zero, we can decompose $ \hat{\U}$ and $\hat{\V}$ into two matrices, i.e., $\hat{\U} = [\U, \tilde{\U}]$ and $\hat{\U} = [\V, \tilde{\V}]$, where $ \U $ and $ \V $ are the matrices of first $ R $ vectors in $\hat{\U} $ and $\hat{\V} $, respectively, $ \tilde{\U} = \mathbf{0} $, and $ \tilde{\V} $ is the matrix of remained $ B-R $ vectors in $\hat{\V} $. Thus we have $ \X = \hat{\U} \hat{\V}^{\textrm{T}} = \U \V^{\textrm{T}}$, and $ \tilde{\U}=\X \tilde{\V}=\mathbf{0} $.  Denote $\X_i$ and $\U_i$ as the $i^{th}$ row vectors of $\X$ and $\U$ ($i=1,2,\cdots, MN$), respectively.

\textbf{Proof of (a)} Suppose $\X(i,:)$, $\X(j,:)$ are two identical vectors, and have different representative coefficient vectors under $ \V $, set as $ \boldsymbol{\alpha} $ and $ \boldsymbol{\beta} $ respectively. Then we have
\begin{equation}
\mathbf{0} = \X(i,:) - \X(j,:) = (\boldsymbol{\alpha} - \boldsymbol{\beta}) \V.
\end{equation}
Since $ \V $ is orthogonal matrix, we have $ \boldsymbol{\alpha}= \boldsymbol{\beta} $.

\textbf{Proof of (b)} For any $1\leq i,j\leq MN$, we have
\begin{equation}
\label{innerPro}
\begin{split}
\X_i\X_j^\textrm{T} &= \hat{\U}_i \hat{\V}^{\textrm{T}} \hat{\V} \hat{\U}_j = \U_i\U_j^{\textrm{T}}, \\
\| \U_i\|_2 &= \| \hat{\U}_i\|_2 = \| \X_i \hat{\V}\|_2=\| \X_i\|_2.\\
\end{split}
\end{equation}
Therefore,
\begin{equation}\label{Angle}
\begin{split}
\frac{\U_i\U_j^{\textrm{T}}}{\|\U_i\|_2^2\|\U_j\|_2^2}= \frac{\X_i\X_j^\textrm{T}}{\|\X_i\|_2^2\|\X_j\|_2^2}
\end{split}
\end{equation}
which means $\mbox{Angle}(\U_i,\U_j)=  \mbox{Angle}(\X_i,\X_j)$. Beside,
\begin{equation}\label{Dist}
\begin{split}
  &\mbox{Dist}(\U_i,\U_j) = \sqrt{(\U_i-\U_j)(\U_i-\U_j)^\textrm{T}}
  \\&= \sqrt{\|\U_i\|_2^2-2\U_i(\U_j)^{\textrm{T}}+\|\U_j\|_2^2}
  \\&= \sqrt{(\X_i-\X_j)(\X_i-\X_j)^\textrm{T}}= \mbox{Dist}(\X_i,\X_j).
\end{split}
\end{equation}
This completes the proof.
\end{proof}	
\end{document}